\def\profilefoundryfinal{}
\def\profilefoundrymatchdesk{}

\newif\ifpfacl
\newif\ifpfpreprint
\newif\ifpffinal
\newif\ifpfsubmission
\ifdefined\profilefoundryacl\pfacltrue\fi
\ifdefined\profilefoundrypreprint\pfpreprinttrue\fi
\ifdefined\profilefoundryarxiv\pfpreprinttrue\fi
\ifdefined\profilefoundryfinal\pffinaltrue\fi
\ifpfpreprint
  \pfsubmissionfalse
\else
  \ifpffinal
    \pfsubmissionfalse
  \else
    \pfsubmissiontrue
  \fi
\fi

\ifpfacl
  \documentclass[11pt]{article}
  \ifpfpreprint
    \usepackage[preprint]{acl}
  \else
    \usepackage[review]{acl}
  \fi
\else
  \documentclass{article}
  \ifpffinal
    \usepackage[final]{colm2026_conference}
  \else
    \ifpfpreprint
      \usepackage[preprint]{colm2026_conference}
    \else
      \usepackage[submission]{colm2026_conference}
    \fi
  \fi
\fi

\usepackage[T1]{fontenc}
\usepackage[utf8]{inputenc}
\ifpfacl
  \usepackage{times}
  \usepackage{latexsym}
\fi
\usepackage{microtype}
\usepackage{graphicx}
\usepackage{booktabs}
\usepackage{tabularx}
\usepackage{array}
\usepackage{multirow}
\usepackage{amsmath}
\usepackage{xcolor}
\usepackage{colortbl}
\ifpfacl\else
  \usepackage{hyperref}
\fi
\usepackage{url}
\ifpfacl\else
  \usepackage{lineno}
\fi
\usepackage{placeins}
\ifpfacl
  \usepackage{stfloats}
\fi

\definecolor{darkblue}{rgb}{0, 0, 0.5}
\ifpfacl\else
  \hypersetup{colorlinks=true, citecolor=darkblue, linkcolor=darkblue, urlcolor=darkblue}
\fi

\definecolor{PFCharcoal}{HTML}{3C3C46}
\definecolor{PFGold}{HTML}{B07D2B}
\definecolor{PFGreen}{HTML}{3F7D4F}
\definecolor{PFMagenta}{HTML}{A8487F}
\definecolor{PFPurple}{HTML}{6B5AA6}
\definecolor{PFRust}{HTML}{A8503A}
\definecolor{PFBlue}{HTML}{3D6FA3}
\definecolor{PFMuted}{HTML}{5A6472}
\definecolor{PFTeal}{HTML}{2F8079}
\definecolor{PFLightGold}{HTML}{F8F1E6}
\definecolor{PFLightGreen}{HTML}{EEF4EF}
\definecolor{PFLightMagenta}{HTML}{F6EEF3}
\definecolor{PFLightPurple}{HTML}{F1EFF7}
\definecolor{PFLightRust}{HTML}{F7EFEA}
\definecolor{PFLightBlue}{HTML}{EDF2F8}
\definecolor{PFLightTeal}{HTML}{EEF5F4}
\definecolor{PFLightNeutral}{HTML}{F6F6F4}

\newcolumntype{Y}{>{\raggedright\arraybackslash}X}
\newcolumntype{P}[1]{>{\raggedright\arraybackslash}p{#1}}
\newcommand{\pf}{\textsc{ProfileFoundry}}
\newcommand{\core}{\texttt{ProfileFoundry-\allowbreak Synthetic-\allowbreak Person-\allowbreak Objects}}
\newcommand{\code}[1]{\texttt{\detokenize{#1}}}

\makeatletter
\newcommand{\pffirstpageavailability}{%
  \begingroup
  \renewcommand{\thefootnote}{}%
  \long\def\@makefntext##1{\parindent 0pt\noindent ##1}%
  \footnotetext{%
    \ifpfsubmission
    The dataset, schema, validation artifacts, and generation code are publicly released.
    Links are withheld during anonymous review and will be included in the camera-ready version.
    \else
    \begin{tabular}{@{}l@{\ }l@{}}
    \textbf{Code:} &
      \href{https://github.com/selvamsriram/ProfileFoundry}{\nolinkurl{github.com/selvamsriram/ProfileFoundry}}\\
    \textbf{Dataset:} &
      \href{https://huggingface.co/datasets/srirxml/ProfileFoundry-Synthetic-Person-Objects}{\nolinkurl{hf.co/datasets/srirxml/ProfileFoundry-Synthetic-Person-Objects}}
    \end{tabular}%
    \fi}%
  \addtocounter{footnote}{-1}%
  \endgroup
}
\makeatother
\newcommand{\pfinlinetablecaption}[2]{%
  \refstepcounter{table}%
  \vspace{0pt}%
  {\scriptsize\centering
  \begin{minipage}[t][1.45\baselineskip][t]{\linewidth}
  \centering Table~\thetable: #1\label{#2}
  \end{minipage}\par}%
}
\title{ProfileFoundry: A Synthetic Person-Object Substrate for \\ Privacy, Memory, and Tool-Use Evaluation in LLM Agents}

\ifpfsubmission
  \ifpfacl
    \author{Anonymous ACL submission}
  \else
    \author{Anonymous authors\\Paper under double-blind review}
  \fi
\else
  \author{%
\makebox[\dimexpr\textwidth-2\tabcolsep\relax][c]{%
\begin{minipage}{0.96\textwidth}
\centering
\begin{tabular}{@{}c@{}}
Sriram Selvam \hspace{3.2em} Anneswa Ghosh\\[0.18em]
{\normalfont\small Independent Researcher}\\[0.12em]
{\normalfont\small
\href{mailto:selvamsriram@gmail.com,anneswaghosh@gmail.com}{%
\texttt{\{selvamsriram,anneswaghosh\}@gmail.com}}}
\end{tabular}
\end{minipage}}}

\fi

\begin{document}
\ifpfsubmission
  \ifpfacl\else
    \linenumbers
  \fi
\fi
\maketitle
\ifpffinal
  \lhead{Published as a workshop paper at HAIPS@COLM 2026}
\fi
\begin{abstract}
Foundation-model research increasingly needs data about people: user state, personal histories, relationships, contact-like fields, documents, and longitudinal updates. Real user data is difficult to share, perturb, audit, or redistribute responsibly, while independently generated fake fields rarely preserve the cross-field and temporal consistency needed for controlled evaluation. We present \pf{}\pffirstpageavailability, a deterministic generator and fixed reference release of 100,000 adult synthetic Person Objects across eight locales. Each object combines a typed current snapshot, household, family, and employer links, snapshot-aligned events, normalized relational views, and generation provenance. The release contains 709,228 events, 40,338 households, 52,491 employers, and 518,564 directed relationship edges. We report evidence in separate categories: selected population-marginal comparisons, per-object invariant checks, release-wide referential and temporal closure, and coincidence/provenance screens.\ifdefined\profilefoundrymatchdesk{} A pilot case study, MatchDesk, uses certified coincidences, typed events, and history truncation to evaluate whether models distinguish corroborated identity evidence from underdetermined matches.\fi{} \pf{} is not a population-fidelity model, a rendered-text corpus, or a formal privacy mechanism. Instead, it is a responsible synthetic source layer for constructing downstream foundation-model evaluations involving memory, privacy, document understanding, record linkage, and agent state while keeping the synthetic person behind each artifact inspectable.
\end{abstract}
\section{Introduction}

Research on language-model privacy, memorization, and user-state behavior has become increasingly dataset-dependent. Memorization studies have used inserted canaries to measure exposure~\citep{carlini2019secret}, web-scale extraction attacks to recover verbatim training snippets containing public PII~\citep{carlini2021extracting}, and real user traces to study privacy inference beyond memorization~\citep{staab2024beyond}. PII detection and redaction work has introduced synthetic span-labeled corpora such as SPY and Nemotron-PII~\citep{savkin2025spy,nvidia2025nemotronpii}. Long-term memory and personalization benchmarks use multi-session conversations, user histories, or curated profiles to test whether models can recall, update, and apply user state~\citep{maharana2024evaluating,wu2025longmemeval,salemi2024lamp,jiang2025personamem}. These lines of work show that synthetic or semi-synthetic personal data is already central to NLP research, especially when real personal data is unsafe or unreleasable.

These datasets leave a common lower-level gap. Canary studies provide controlled secrets, but not coherent people. Extraction and inference studies often depend on real web or social traces, which are difficult to redistribute, perturb, or audit as synthetic identities. PII corpora provide labeled spans in rendered text, but usually do not expose the underlying person graph that generated those spans. Memory and personalization benchmarks provide fixed conversations, histories, or user profiles for a particular evaluation, but not a scalable population from which new privacy, retrieval, dialogue, document, or agent-state datasets can be derived under the same schema and seed. The missing artifact is therefore not another single task benchmark; it is a reusable source layer: an auditable population of internally consistent synthetic people whose identifiers, relationships, employers, addresses, timelines, and provenance can be rendered into many downstream study designs.

\begin{figure*}[t]
  \centering
  \includegraphics[width=0.99\textwidth]{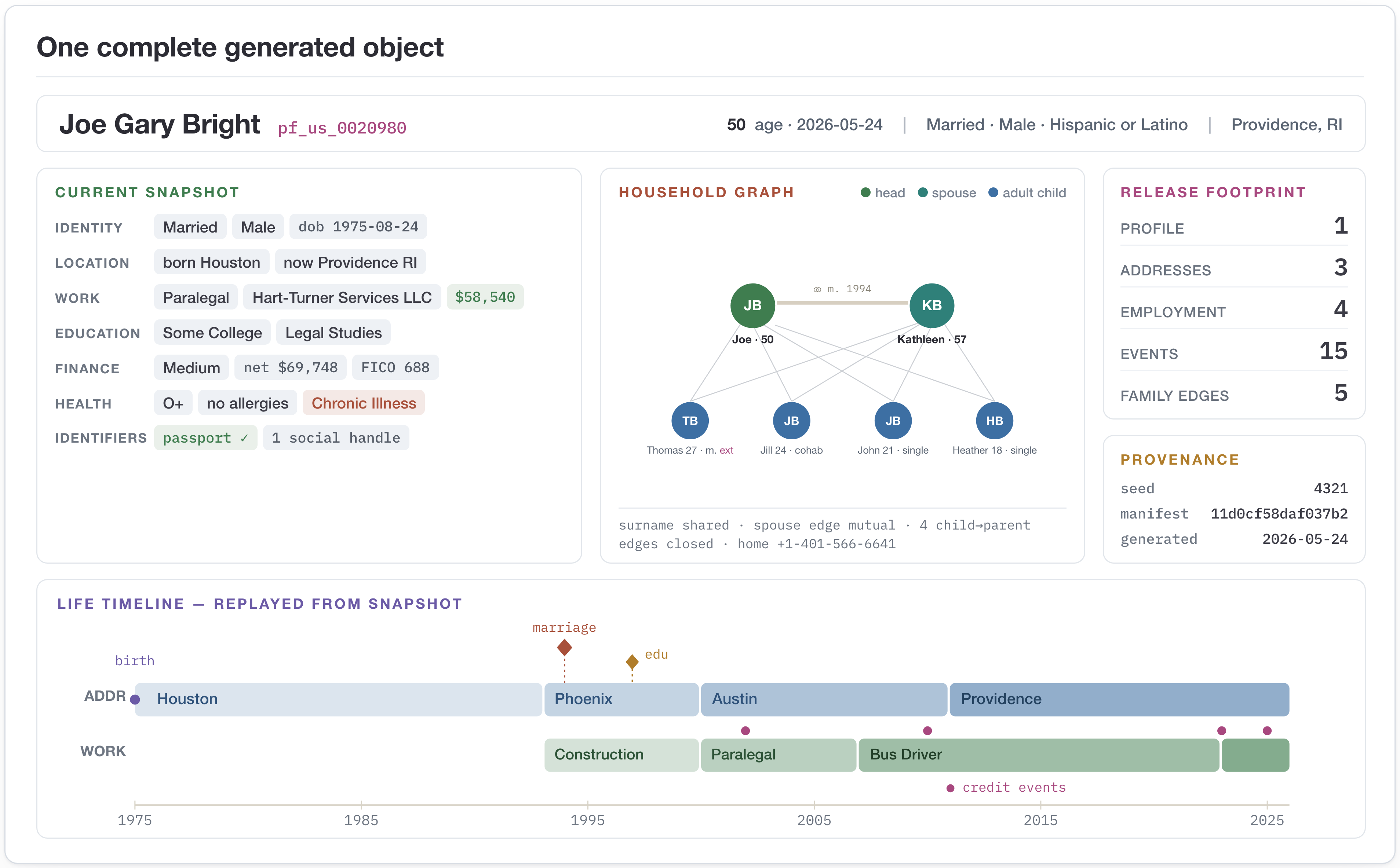}
  \caption{A sample released person object from en-US locale.}
  \label{fig:complete_object}
\end{figure*}

\pf{} is designed as this base layer. It provides structured adult Person Objects whose demographics, contacts, households, family links, employers, addresses, identifiers, and event histories are generated and validated together. Researchers can transform these objects into task-specific artifacts such as PII-laced text, memory entries, retrieval documents, forms, dialogue states, linkage pairs, perturbation sets, or exposure corpora while retaining a known source object and deterministic generation path. Existing resources support important pieces of this workflow, but they generally expose different layers: rendered text, task instances, learned or simulated tables, local fake fields, or domain-specific longitudinal records.

\pf{} does not claim that structured synthetic profiles are new. Its distinct contribution is to separate and release a broader source artifact: a versioned schema, represented-person and employer graph, snapshot-aligned events, normalized analytical views, deterministic generator, and release-level evidence. Appendix Tables~\ref{tab:comparison-text-memory} and~\ref{tab:comparison-population} record the artifact-level distinction.

\pf{} is especially relevant now because language systems increasingly act through tools, memory, retrieval, and persistent user state. A source object makes controlled confounders possible: two people can share a household, employer, surname, or city without being the same person; an earlier address or job can be superseded by a later event; and rendered spans can retain links to the fields and events that produced them. The present paper establishes the source layer and its released evidence, not downstream model effectiveness.
\ifdefined\profilefoundrymatchdesk
Section~\ref{sec:matchdesk} uses that source layer in a pilot evidence-sufficiency case study; it is not presented as a completed downstream benchmark or a general ranking of model effectiveness.
\fi

Our contributions are:
\begin{itemize}
    \item \textbf{Person Object abstraction:} a reusable Person Object with snapshot fields, graph links, typed events, reserved document hooks, and generation provenance for constructing sensitive-data-like foundation-model evaluations without releasing real user traces.
    \item \textbf{Internally consistent linked generation:} the generator first commits to household roles, represented relationships, shared addresses, employers, and snapshot facts, then derives person records, graph edges, foreign keys, and snapshot-aligned events from those commitments so linked fields simulate real person.
    \item \textbf{Executable generator and SDK:} a Python package and CLI for deterministic profile and household generation, scaled release builds, validation, export, and release-rebuild workflows.
    \item \textbf{Audited reference release:} the 100K release includes canonical JSONL, a complete viewer Parquet file, flat snapshots, normalized relational tables, manifest hashes, and a dataset card, accompanied by validation, leakage, and report-quality evidence.
    \item \textbf{Artifact-accountability protocol:} a release audit that separates distributional gaps, declared consistency, referential and temporal closure, coincidence screens, reserved-domain email checks, provenance, and documentation-drift evidence.
\end{itemize}
\section{Related Work}
\label{sec:related-work}

Synthetic personal data spans privacy, personalization, agent evaluation, record linkage, statistical disclosure control, and domain simulation, but these areas usually release different artifacts. Privacy corpora expose rendered text and labels; memory benchmarks expose fixed histories or tasks; population and tabular systems expose linked records, learned relationships, or domain simulations; and fake-data libraries expose localized fields or schema-generated records. \pf{} targets the reusable source layer: schema-governed Person Objects with inspectable links, typed state-changing events, normalized exports, provenance, and release-level evidence.

\paragraph{Privacy-rich text and PII corpora.}
Language-model privacy work shows that models can memorize sensitive strings and infer private attributes without verbatim reproduction~\citep{carlini2019secret,carlini2021extracting,staab2024beyond}. Privasis, PANORAMA, SynthPAI, SPY, Nemotron-PII, Gretel's multilingual financial PII data, and PIIBench address releasable private text, PII/PHI detection, span labeling, de-identification, memorization, or corpus unification~\citep{kim2026privasis,selvam2025panorama,yukhymenko2024synthetic,savkin2025spy,nvidia2025nemotronpii,gretel2024syntheticpii,jha2026piibench}. Several use profiles internally and provide substantial human or benchmark evaluation. \pf{} differs at the exposed layer: it releases the structured identities, households, employers, relationships, identifiers, events, normalized views, and provenance from which text corpora can be derived, but it does not itself release rendered prose or span labels.

\paragraph{Personalization, memory, and private-user benchmarks.}
PersonaBench, LaMP, LoCoMo, LongMemEval, and PersonaMem evaluate personal-information QA, personalized tasks, long-term conversational memory, temporal reasoning, knowledge updates, or user-aware response generation~\citep{tan2025personabench,salemi2024lamp,maharana2024evaluating,wu2025longmemeval,jiang2025personamem}. LoCoMo and PersonaMem have meaningful temporal structure, and PersonaBench uses a social graph during construction. Their primary interfaces remain fixed conversations, documents, or tasks. \pf{} instead makes source state reusable through stable profile IDs, represented-person links, employer IDs, relationship edges, and typed histories.

\paragraph{Personas and simulated behavior.}
PersonaChat, Persona Hub, and generative-agent systems use persona facts, descriptions, memories, reflection, planning, or social behavior as conditioning signals~\citep{zhang2018personalizing,ge2024personaHub,park2023generative}. Generative Agents in particular model social behavior over simulated time. \pf{} is narrower in behavior and broader in release structure: it provides inspectable personal-state objects and links rather than a behavioral simulation or persona-prompt collection.

\paragraph{Synthetic populations, record linkage, and tabular synthesis.}
Pseudopeople exposes stable simulant, household, and employer identifiers across simulated administrative records and is the closest population-level comparator~\citep{haddock2024pseudopeople,pseudopeople2026docs}. Synthea provides longitudinal, linked patient records within healthcare~\citep{walonoski2018synthea}. Febrl and GeCo support generation, corruption, and linkage workflows~\citep{christen2008febrl,tran2013geco}. synthpop, SDV, PrivBayes, and PrivSyn provide statistical, relational, sequential, or differentially private synthesis~\citep{nowok2016synthpop,patki2016sdv,zhang2017privbayes,zhang2021privsyn}. \pf{} does not replace these systems or claim broader locale fidelity; it packages an NLP-facing person-state layer across eight configured locales, with canonical objects, normalized views, represented links, typed events, and release evidence. The marginal audit covers the US, UK, IN, CA, and AU locales, while IE, NZ, and PH remain lightly validated.

\paragraph{From fake fields to coupled objects.}
Faker generates localized values and composite profiles, while current Mimesis schemas can express foreign-key references between generated schemas~\citep{faker2025,mimesis2026schema}. \pf{} uses fake-data providers at the leaf level but adds release-specific cross-field constraints, represented household and employer commitments, snapshot-aligned histories, deterministic identifiers, and audit artifacts. Its novelty is therefore the combination and exposure of these layers, not the first generation of synthetic names, profiles, households, or longitudinal records.

Appendix Tables~\ref{tab:adjacent-resources}, \ref{tab:comparison-text-memory}, and~\ref{tab:comparison-population} summarize the closest adjacent resources and give the resource-by-resource evidence rubric.
\section{Object Contract}

\pf{} generates from a constrained object space. A Person Object is not a bag of independent fake fields; it is a typed adult record whose snapshot fields, household references, employer links, event history, normalized rows, and provenance are generated as mutually constrained commitments. Figure~\ref{fig:complete_object} shows the object at the level a downstream NLP system would consume: current fields, linked household members, snapshot-aligned address and job histories, release rows, and seed and manifest metadata.

The canonical schema has four surfaces. The \emph{snapshot} contains identity, contact, addresses, employment, education, finance, health, government IDs, household ID, family graph, events, reserved document hooks, and generation metadata. The \emph{graph} surface contains household membership, spouse or partner links, parent--adult-child links, sibling links, colleague links, and employer IDs. The \emph{temporal} surface contains typed events: \code{birth}, \code{education}, \code{move}, \code{job_change}, \code{marriage}, \code{divorce}, \code{name_change}, and \code{credit_event}. The \emph{provenance} surface records global seed, profile seed, SDK version, generation date, exported timestamp, and reference manifest hash.

The schema is also an interoperability contract. The in-memory source is implemented with Pydantic models and exported as JSON Schema for non-Python consumers~\citep{pydantic}. Canonical JSONL preserves complete nested objects, while Parquet views expose the same source record as row-counted relational tables. This separation lets one source object seed an agent-memory store, rendered document, PII-tagged passage, linkage pair, or perturbation set without discarding provenance. Appendix Table~\ref{tab:schema_coverage} maps each schema group to the release evidence that supports it.
\vspace{-1em}
\section{Constrained Generation}

\subsection{Household-First Linkage}

The generator implements the object contract as the cascade summarized in Appendix Figure~\ref{fig:cascade}. It first samples a household plan, turns that plan into member hints, materializes each person under those hints, closes links, constructs events from the finalized snapshot, and then exports audited release files. This ordering creates the conditions under which marital pins, adult-child slots, shared surnames, shared addresses, family edges, employer reuse, historical rows, and temporal constraints can be made coherent before release validation.

Generation begins with a locale-specific household composition table. For en-US, the eight implemented weights are single households at 0.285, couples without children at 0.265, couples with represented adult children at 0.190, single-parent households with represented adult children at 0.090, cohabiting households without children at 0.050, cohabiting households with represented adult children at 0.030, multigenerational households at 0.045, and unrelated-adult households at 0.045. The weights sum to one. The ``child'' slots in v1.0 are adult children living with parents, normally age 18--28.

\begin{figure*}[t]
  \centering
  \includegraphics[width=0.99\textwidth]{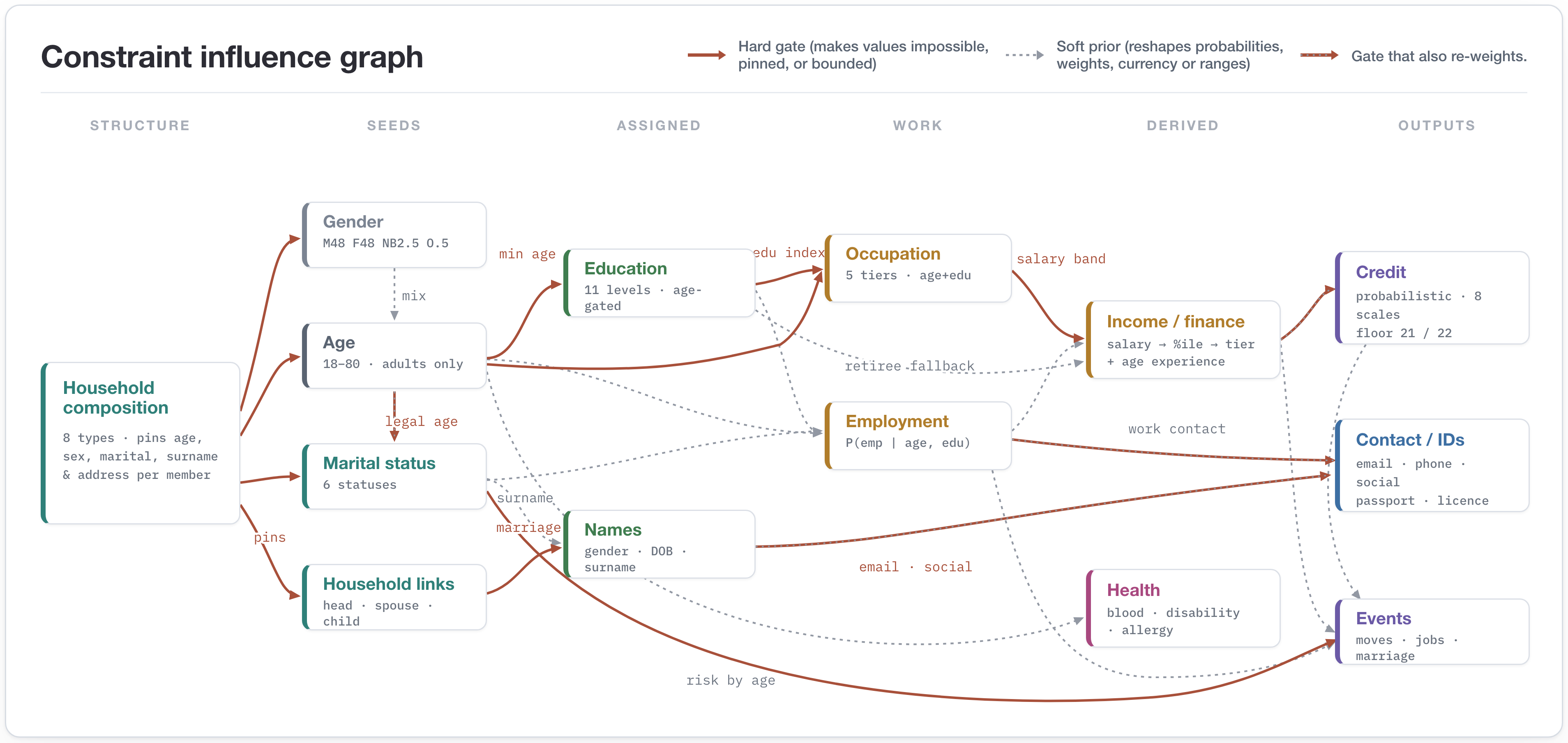}
  \caption{Constraint influence graph, the full dependency map: which factor constrains which, and whether the effect is a hard gate, a soft prior, or both.}
  \label{fig:constraint_influence_graph}
\end{figure*}

Figure~\ref{fig:constraint_influence_graph} shows the dependency map behind this household-first design and why it is not equivalent to sampling marital status and then adding dependents. Composition determines which role slots exist. Role slots then carry sex hints, age bands, marital pins, surname rules, and shared-address context into the person factory. The family builder closes the represented graph: head-spouse mutual links, cohabiting partner links, parent-child links from head and spouse to every adult-child slot, sibling links among adult children, and grandparent-to-head links in multigenerational households. When a married or cohabiting partner is not represented, the partner field is explicitly marked external rather than silently omitted or assigned a dangling profile ID.

Several small constraints are deliberately visible in the generated object. Spouse ages are sampled within an 11-year band around the head with mass near small gaps; adult-child ages are capped by the youngest represented parent; adult siblings are spaced by at least three current-age years when possible; co-residents share a home phone when one is present; and emergency contacts prefer spouse, parent, sibling, then another household adult. Current employers are assigned from a deterministic locale pool of 250 employers; compatible working adults in a household share an employer with probability 0.25. These rules support internal coherence but are modeling heuristics, not claims about every locale's real household structure. Appendix Figure~\ref{fig:employer_network} shows the resulting employer and colleague-edge structure in the release.

\subsection{Field Materialization and Partial Replay}

Within each member slot, fields are generated in dependency order. Age is sampled first and gates downstream choices. Education zeroes out levels whose minimum completion age exceeds the sampled age: Bachelor requires age 21, Master 22, and Doctorate 25. Marital status is sampled by age band and sex, then adjusted by feasibility rules: below the configured modeling floor the state is Single; former-partner states are impossible under 19; widowhood under 40 is multiplied by 0.05; cohabiting is multiplied by 0.40 for ages 35--44 and by 0.10 for ages 45--54; and cohabiting is zero at 55+. These are generator assumptions rather than legal or demographic ground truth. Appendix Figures~\ref{fig:age_atlas} and~\ref{fig:education_career} provide the detailed visual atlas.

Occupation uses hand-authored education- and age-gated title weights. At age 30, High School profiles favor Entry/Service and Skilled/Technical titles; Bachelor profiles favor Professional/Analyst and Management/Executive titles. Master and Doctorate profiles exclude Entry/Service but retain small-weight overqualification paths; Appendix Figure~\ref{fig:outlier_policy_designer} shows blocked, bent, and weighted outliers. Title tier bounds salary using bands aligned only to U.S. BLS ranges~\citep{bls2024oews}; salary percentile drives finance tier; credit uses locale-specific bureau scales. The 300--850 US FICO centers are 480, 580, 680, 740, and 790; non-US scale labels are explicit. Unlike sourced age-by-sex, education, and marital-status marginals, occupation weights, finance thresholds, and credit modeling lack external calibration. Sex/gender is not an input to occupation, salary, income, or credit. These heuristics can encode socioeconomic assumptions documented as limitations.

After link closure, temporal history is constructed from the finalized snapshot rather than sampled independently. This avoids stale timeline rows caused by linkages or current fields changing after draft events were sampled. The event backfill always includes birth, appends the current move when an address exists, appends the current job-change when employment exists, aligns represented-spouse marriage dates, emits name-change and divorce events when the snapshot implies them, and samples prior moves or jobs only when enough lifetime span exists. Validation checks that the latest covered move and job events agree with current address and employment, that address rows retain source-event IDs, and that events do not predate DOB. This is snapshot-aligned history with partial replay over declared fields, not complete event sourcing of the whole Person Object. Appendix Figure~\ref{fig:temporal_replay_designer} shows a concrete example.

Contact and ID fields are also downstream of earlier commitments. Work email depends on the final employer name and employer ID; phone formatting depends on locale and address region; personal and work emails use reserved \code{profilefoundry.example} domains following RFC~2606~\citep{rfc2606}. Social handles are age-gated, LinkedIn requires age 16, banned-platform rules are locale-aware, and platform selection changes with age rather than being flat across the adult range.
\par\vspace{-0.95em}
\section{Release}
\label{sec:release}
\vspace{-0.5em}

We distribute \pf{} in two complementary forms: an executable Python package and a fixed 100K reference dataset. The package supports incremental fixes and task-specific generation under the same schema; \core{} provides a stable, citable population without requiring users to rerun generation. Hosted access conditions and package metadata are governed by the accompanying artifacts, so the paper does not depend on an unrestricted-hosting claim.

\vspace{-0.8em}
\subsection{Python Package Release}
\vspace{-0.45em}

The package can be installed with \code{pip install profilefoundry}. It emits the Person Object schema described above, so a generated profile can be consumed as nested JSON, converted into normalized tables, or used as a seed object for downstream NLP artifacts. Table~\ref{tab:release_package_surface} summarizes the command guide.

\subsection{100K Reference Release}

\core{} is the fixed reference artifact for direct use and comparison. It is intended for users who want a stable population without rerunning the generator, while the package supports task-specific generation under the same object contract. Table~\ref{tab:release_reference_metadata} lists the release contents and reproducibility pins.

The release is deliberately more than a flat profile table. The canonical JSONL preserves nested Person Objects, while the Parquet views expose scalar, temporal, relational, household, employer, education, address, social-handle, and allergy views for downstream analysis. The 14-file local bundle contains 709{,}228 events, 518{,}564 directed relationships, 167{,}089 addresses, 111{,}955 employment rows, 74{,}738 education rows, 40{,}338 households, and 52{,}491 employers. Appendix Figure~\ref{fig:release_topology} maps the release inventory and object topology, Appendix Table~\ref{tab:storage_appendix} gives the complete row-counted inventory, and Appendix Figure~\ref{fig:profile_coverage} reports profile-level coverage.

The reference bundle and associated reports are rebuilt through the repository release workflow, "\code{python scripts/run_full_core.py --generation-date 2026-05-24 --exported-at 2026-05-24 --skip-hibp}".
The fixed data identity is pinned by its manifest identifier, per-file hashes, row counts, seed, and dates. Appendix Figure~\ref{fig:reproducibility} and Table~\ref{tab:repro_checklist} record the reproducibility pin and the command-level verification checklist.

\ifpfsubmission\nolinenumbers\fi
\ifpfacl
\begin{table*}[!t]
\fi
\vspace{-0.35em}
\noindent\begingroup
\centering
\setlength{\tabcolsep}{1.8pt}
\begin{minipage}[t]{0.57\linewidth}
\vspace{0pt}
\fontsize{6.3}{6.7}\selectfont
\renewcommand{\arraystretch}{0.90}
\begin{tabularx}{\linewidth}{@{}P{0.33\linewidth}Y@{}}
\toprule
Command & Release role \\
\midrule
\code{profilefoundry verify} & One-profile-per-locale smoke test. \\
\code{profilefoundry person} & Generates one Person Object. \\
\code{profilefoundry household} & Linked household with shared address, family, partner, and emergency-contact constraints. \\
\begin{tabular}[t]{@{}l@{}}\code{profilefoundry}\\\code{scale-households}\end{tabular} & Generates household-preserving JSONL at scale. \\
\code{profilefoundry validate} & Marginal, consistency, replay, collision, Wikidata, and email checks. \\
\code{profilefoundry export} & JSONL, Parquet, dataset-card, manifest, validation, and leakage outputs. \\
\bottomrule
\end{tabularx}
\pfinlinetablecaption{CLI commands.}{tab:release_package_surface}
\end{minipage}\hfill
\begin{minipage}[t]{0.39\linewidth}
\vspace{0pt}
\fontsize{6.3}{6.7}\selectfont
\renewcommand{\arraystretch}{1.07}
\begin{tabularx}{\linewidth}{@{}P{0.43\linewidth}Y@{}}
\toprule
Release field & Value \\
\midrule
Profiles & 100{,}000 adults \\
Locales & US 35K; UK/IN 20K; CA/AU 8K/7K; IE/NZ/PH 4K/3K/3K \\
Validation split & Full: US, UK, IN, CA, AU (90\%); lighter/excluded: IE, NZ, PH \\
Surfaces & canonical JSONL plus 11 Parquet views \\
\bottomrule
\end{tabularx}
\pfinlinetablecaption{Release contents.}{tab:release_reference_metadata}
\end{minipage}
\par\endgroup
\vspace{-0.85em}
\ifpfacl
\end{table*}
\fi
\ifpfsubmission\linenumbers\fi
\par\vspace{-0.75em}
\section{Audit}
\label{sec:audit}
\vspace{-0.4em}

The \core{} audit asks whether the reference release is usable as a synthetic person-object substrate: whether its limited population comparisons are disclosed, its declared references and covered histories resolve, its objects satisfy the implemented rules, its coincidence risks are screened, and its reports remain tied to the artifact. These forms of evidence answer different questions and are not combined into one quality score.

\subsection{Population Fit and Declared Consistency}

For the five full-validation locales (US, UK, IN, CA, and AU), the validator compares generated age-by-sex, education, and marital-status bucket shares with public reference tables. For each marginal, it reports the largest absolute bucket-share difference, an $L_{\infty}$ marginal gap rather than a Kolmogorov--Smirnov statistic. Separately, it checks whether every generated object in those locales satisfies the declared structural and covered replay invariants. IE, NZ, and PH are included in the release but excluded from the locked marginal-fit table because their reference coverage is lighter.

The locked targets were maximum gap $\le 0.10$ per attribute and mean gap $\le 0.07$ per locale. The release does not meet the mean target, and we report the miss directly: locale means range from 0.074 to 0.089, with IN largest because its male and female age gaps are both approximately 0.124. In contrast, all 90{,}000 profiles in the five full-validation locales pass the declared consistency suite. This pass establishes agreement with implemented constraints; it does not independently validate the realism of those constraints, joint distributions, household topology, or event rates.

\subsection{Object, Linkage, and Temporal Closure}

The object audit covers age gates, address validity after DOB, phone and locale rules, reserved-domain contacts, identifier uniqueness, employer foreign keys, relationship endpoints, household closure, whole-household selection, and partial replay. Across the full 100K normalized release, relationship source and target endpoints have zero misses; employment rows and current profile employer references have zero missing employer foreign keys; household member counts sum to 100{,}000; and represented spouse links are mutual in 49{,}072 of 49{,}072 cases. Parent--child, partner, sibling, household-member, and colleague reciprocal commitments have zero reported reverse-edge misses. External spouse references are explicit sentinel cases rather than broken edges.

Temporal checks cover 709{,}228 typed events. All 167{,}089 address rows retain source-event identifiers, every profile has one current address, and no event predates date of birth. Latest covered move and job-change events agree with current address and employment. These checks establish source linkage and partial replay for covered fields; they do not establish complete event sourcing or realism of real-world transition rates.
\par\vspace{-0.65em}
\subsection{Coincidence, Collision, and Drift Screens}

We screen coincidence and drift separately from privacy claims: 7 repeated name+DOB tuples, 1{,}038 repeated name+birth-city tuples, 0 personal-email self-collisions, 0 email-syntax findings, and 342 Wikidata Bloom flags. The Bloom filter covers 683{,}897 humans with known birth dates and at least five sitelinks, indexing \code{name|birth_year} and \code{name|birth_city} keys at target false-positive rate $10^{-4}$~\citep{vrandevcic2014wikidata}. Reserved \code{profilefoundry.example} domains make email evidence syntax/uniqueness-only; a report-quality verifier checks manuscript counts, figures, validation, leakage, and manifest metadata for drift. Appendix Figures~\ref{fig:claim_evidence_ledger}, \ref{fig:household_graph_fig}, and~\ref{fig:audit_attachment_designer}--\ref{fig:leakage_screening} and Appendix Tables~\ref{tab:artifact_evidence}, \ref{tab:household_graph}, \ref{tab:event_inventory}, \ref{tab:validation_interpretation}--\ref{tab:invariants}, and~\ref{tab:reference_sources} give the supporting ledger, closure, inventory, leakage, and provenance evidence.
\par\vspace{-0.55em}
\section{Downstream Use as an NLP Substrate}
\label{sec:downstream}

\pf{} supports downstream datasets by rendering audited fields, links, and events into text or records; intervening through masking, corruption, updates, withholding, or temporal shifts; and evaluating outputs against canonical profile, field, relationship, and event IDs. This covers document understanding, memory and agent-state evaluation, privacy and PII rendering, and record linkage while preserving deterministic provenance and controlled near-misses: shared households, employers, cities, or surnames without identity; stale facts superseded by later events; and graph-grounded ambiguous pairs. Appendix Table~\ref{tab:uses_appendix} summarizes recommended, caveated, and discouraged uses.
\ifdefined\profilefoundrymatchdesk
\subsection{Case Study: Evidence Sufficiency in Screening-Alert Adjudication (MatchDesk)}
\label{sec:matchdesk}

To demonstrate an evaluation enabled by \pf{}'s linked object structure, we instantiate \textsc{MatchDesk}, a screening-alert adjudication harness. Given a rendered customer dossier and an external alert candidate, a model returns \textsc{Match}, \textsc{No\_Match}, or \textsc{Escalate}. The question is deliberately narrower than end-to-end compliance: can a model distinguish corroborated identity evidence from coincidental overlap and defer when the displayed record underdetermines identity? The task is motivated by name-only false-positive screening disputes exemplified by \emph{TransUnion LLC v.\ Ramirez}~\citep{ramirez2021}; real adjudication records are difficult to release because both the records and their resolutions contain PII.

\pf{} provides three controls for this study. Its audited graph and collision structure yield hard negatives, including co-resident relatives sharing a surname and address and the 150-pair \textsc{N-NameDOB} condition: 7 name+DOB pairs plus 143 evaluation pairs sampled from the release-wide set of 1{,}038 repeated name+birth-city tuples. Typed events support same-person positives rendered at an earlier date, making moves, job changes, and former names controlled record-vintage variables. Truncated histories create evidence-insufficient cases without changing the underlying identity truth. Source profile IDs establish latent same/different-person truth, and a deterministic four-rule policy combines that truth with rendered conflicts and corroborating fields. Reapplying the policy to all 3{,}575 frozen pairs produced zero label mismatches. Independent Faker and Mimesis draws~\citep{faker2025,mimesis2025} do not natively provide this combination of identity truth, graph-grounded confusability, and replayable events; Pseudopeople~\citep{haddock2024pseudopeople} provides household and linkage truth over simulated US administrative records, but not the typed person-state histories and rendered dossiers used here.

Table~\ref{tab:matchdesk-stage1} reports results for 540 fixed pairs per model at temperature 0: 1{,}620 calls with 100\% parse success and no refusals. Complete-evidence accuracy is 0.95--1.00. Differences emerge when evidence is insufficient: correct \textsc{Escalate} rates are 0.925, 0.98, and 0.61, while unsupported \textsc{Match} rates are 0.075, 0.006, and 0.381. On a decidable incomplete-history control, over-escalation rates are 0.800, 0.375, and 0.000. These pilot results do not rank general model quality or establish deployment readiness; they show that \pf{} can isolate evidence-sufficiency failures concealed by aggregate complete-evidence accuracy. The frozen pairs, scored rows, and verification script are released under \code{MatchDesk/}.

\par\vspace{-0.4em}
\begingroup
\centering
\footnotesize
\renewcommand{\arraystretch}{0.85}
\setlength{\tabcolsep}{4.5pt}
\begin{tabular}{@{}lccc@{}}
\toprule
Stage-1 metric & gpt-5.4 & grok-4.3 & Kimi-K2.6 \\
\midrule
Core negatives ($n{=}120$) & 1.00 & 1.00 & 1.00 \\
Core positives ($n{=}100$) & 0.95 & 1.00 & 1.00 \\
Name noise L1--L3 ($n{=}90$) & 0.96 & 1.00 & 1.00 \\
Line-up ($n{=}30$) & 1.00 & 1.00 & 1.00 \\
\midrule
Correct escalation ($n{=}160$) & 0.925 & 0.98 & 0.61 \\
Unsupported \textsc{Match} ($n{=}160$) & 0.075 & 0.006 & 0.381 \\
Over-escalation on control ($n{=}40$) & 0.800 & 0.375 & 0.000 \\
\bottomrule
\end{tabular}

\pfinlinetablecaption{\textbf{MatchDesk Stage-1 results.} 540 pairs/model; master seed 23.}{tab:matchdesk-stage1}
\endgroup
\vspace{-0.25em}
\FloatBarrier
\fi
\par\vspace{-0.75em}
\section{Conclusion}

\pf{} argues for a different unit of synthetic personal data: not isolated fake fields, fixed personas, or unreleasable real traces, but schema-governed people whose identities, households, links, histories, exports, and provenance can be inspected together. This matters for stateful NLP because memory, privacy, document, agent, and linkage evaluations often depend on the same hidden requirement: a coherent source person that can be rendered, perturbed, partially replayed, and audited.

The v1.0 release makes that substrate concrete through an executable generator and a 100K reference population with normalized views, manifest hashes, validation reports, closure checks, coincidence screens, and reproducibility commands. The release is intentionally not presented as a perfect population model, a formal privacy mechanism, or a completed downstream benchmark. Its contribution is a reusable, accountable baseline for building evaluations in which the synthetic person behind each artifact remains visible.
\section{Limitations}

\pf{} v1.0 is English-only, adult-only, and limited to eight locales; Faker and project-specific references constrain field coverage and cultural fidelity. Binary sex/gender conditioning, surname sharing, household composition, education--occupation mappings, salary/credit heuristics, disability/health categories, and social-platform rules are simplifying assumptions that may be narrow or stereotyped; Appendix Table~\ref{tab:assumption-ledger} lists risks and disclosures. Household ``children'' are represented adults, so pediatric, school, custody, guardian-consent, and child-safety workflows are unsupported; family links are household-local with external-partner sentinels, limiting extended kin and non-household graphs. Several full-validation locales miss marginal targets; the audit covers univariate age-by-sex, education, and marital-status marginals, not joint distributions, household composition, age gaps, graph degrees, employer structure, or event rates. Invariants test implemented rules, not realism; temporal history is snapshot-backfilled partial replay. Beyond the MatchDesk pilot, the paper includes no completed downstream benchmark, human study, or household-first ablation; finance/health/ID/salary remain synthetic plausibility fields, and package/card/report/manifest metadata require synchronization.
\section{Ethical Considerations}

Synthetic person objects can be misused for impersonation, fraud rehearsal, spam, credential testing, or misleading demonstrations. Reserved email domains, exclusion of minors, provenance records, and collision/notable-person screens do not provide differential privacy, proof of non-resemblance, or permission to use contact-like fields outside controlled research. Non-notable-person resemblance is out of scope because no comprehensive, responsibly accessible registry exists; building one would create privacy risks and substantial acquisition and computation costs. Generated phone numbers, government identifiers, addresses, names, and other identity-like values must not be used to contact, authenticate, evaluate, or make decisions about real people.

Because demographic and socioeconomic rules may reproduce stereotypes, downstream studies should report fields and locales, preserve synthetic labeling and provenance, and audit group-conditioned outcomes. Such results may reflect generator assumptions rather than real-world disparities and are not population evidence. Derived text or documents should retain the dataset card, license, and relevant risk disclosures. \pf{} should not be used to train or validate consequential decision systems about real people.
\section{Acknowledgments}

We gratefully acknowledge the open-source and open-science infrastructure that made \pf{} practical. Faker provides important localized synthetic-data providers used at the leaf-field level, while Pydantic and JSON Schema support the typed object contract and public schema export. We also thank the maintainers of the scientific Python and columnar-data ecosystem, especially NumPy, pandas, and PyArrow, as well as Hugging Face tooling and the \LaTeX{} venue templates used to package and disseminate the release.
\ifpfacl\else
\bibliographystyle{colm2026_conference}
\fi
\bibliography{references}
\clearpage
\appendix
\section{Supplementary Evidence and Reference Material}
\label{sec:appendix}
\suppressfloats[t]
\renewcommand{\topfraction}{0.95}
\renewcommand{\bottomfraction}{0.90}
\renewcommand{\textfraction}{0.05}
\renewcommand{\floatpagefraction}{0.80}
\renewcommand{\dbltopfraction}{0.95}
\renewcommand{\dblfloatpagefraction}{0.80}
\setlength{\floatsep}{7pt plus 2pt minus 2pt}
\setlength{\textfloatsep}{8pt plus 2pt minus 2pt}
\setlength{\intextsep}{7pt plus 2pt minus 2pt}
\setlength{\dblfloatsep}{7pt plus 2pt minus 2pt}
\setlength{\dbltextfloatsep}{8pt plus 2pt minus 2pt}
\makeatletter
\setlength{\@fptop}{0pt}
\setlength{\@fpsep}{8pt plus 1fil}
\setlength{\@fpbot}{0pt plus 1fil}
\setlength{\@dblfptop}{0pt}
\setlength{\@dblfpsep}{8pt plus 1fil}
\setlength{\@dblfpbot}{0pt plus 1fil}
\makeatother

This appendix collects the supporting evidence for \pf{} and follows four buckets. Appendix~\ref{app:generation} documents the \textbf{generation logic}: the age-gated constraint atlas, the education--career--finance signature, the outlier policy, snapshot-aligned partial replay, and modeling assumptions. Appendix~\ref{app:release} describes the \textbf{released artifacts}, covering the Python package command surface, the 100K reference set, its inventory and object topology, per-profile coverage, household and employer graph structure, temporal surface, schema, and use guidance. Appendix~\ref{app:audit} gives the \textbf{audit} evidence: validation against public marginals, reference-data provenance, referential and temporal closure, leakage and collision screening, invariant families, claim-to-evidence mapping, and reproducibility. Appendix~\ref{app:related} provides the resource-by-resource \textbf{related-work comparison}. Counts are descriptive of the fixed reference release unless a caption states otherwise. Internal consistency, distributional fit, coincidence screening, formal privacy, and downstream utility remain separate forms of evidence.

\subsection{Generation Logic}
\label{app:generation}

These figures expand the constrained-generation mechanics summarized in the main text. Age is the master gate that conditions education, marital state, occupation, and finance; the generator then preserves plausible rare combinations rather than collapsing every profile toward the modal path, and finally reconstructs history backward from the finalized snapshot.

\begin{figure*}[!htbp]
  \centering
  \includegraphics[width=\textwidth]{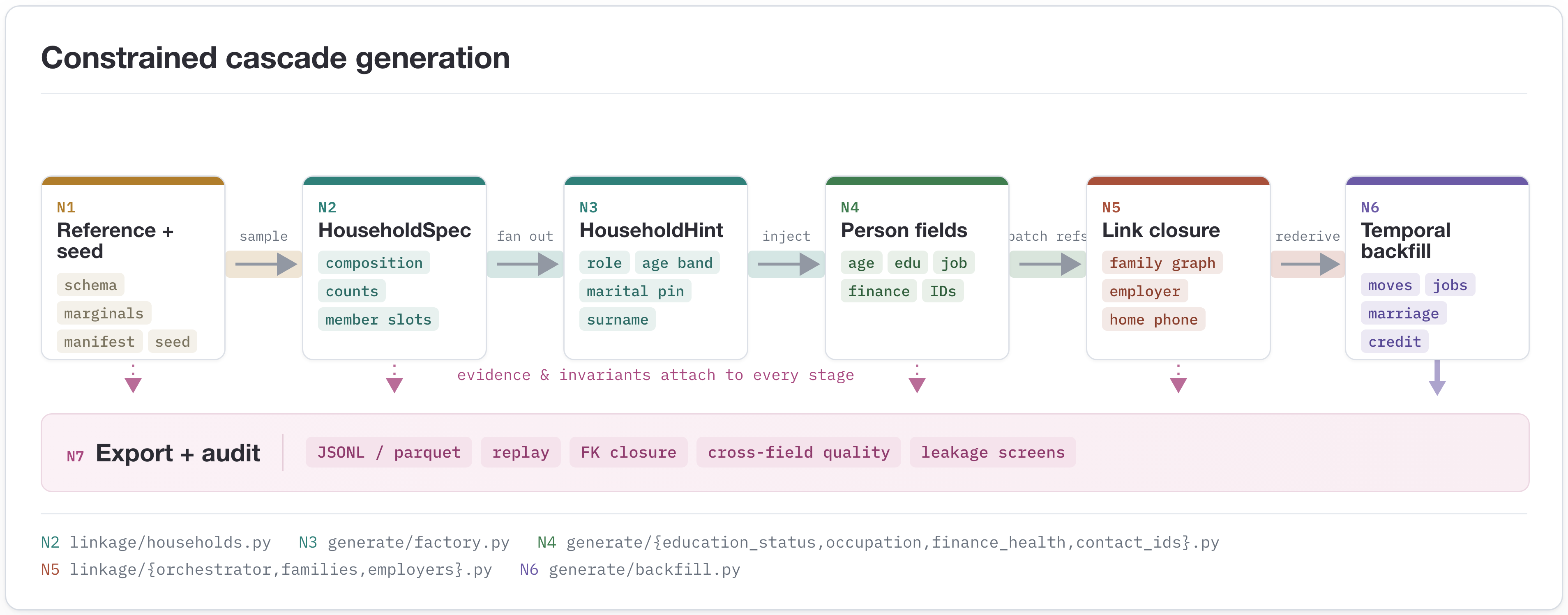}
  \caption{Constrained cascade generation. \pf{} carries constraints forward from reference tables and household plans into person fields, represented-link closure, snapshot-aligned temporal backfill, and export-time evidence checks. Household-first generation is an engineering design choice; without an ablation, this paper does not claim causal superiority over every alternative.}
  \label{fig:cascade}
\end{figure*}

\begin{figure*}[!htbp]
  \centering
  \includegraphics[width=0.88\textwidth]{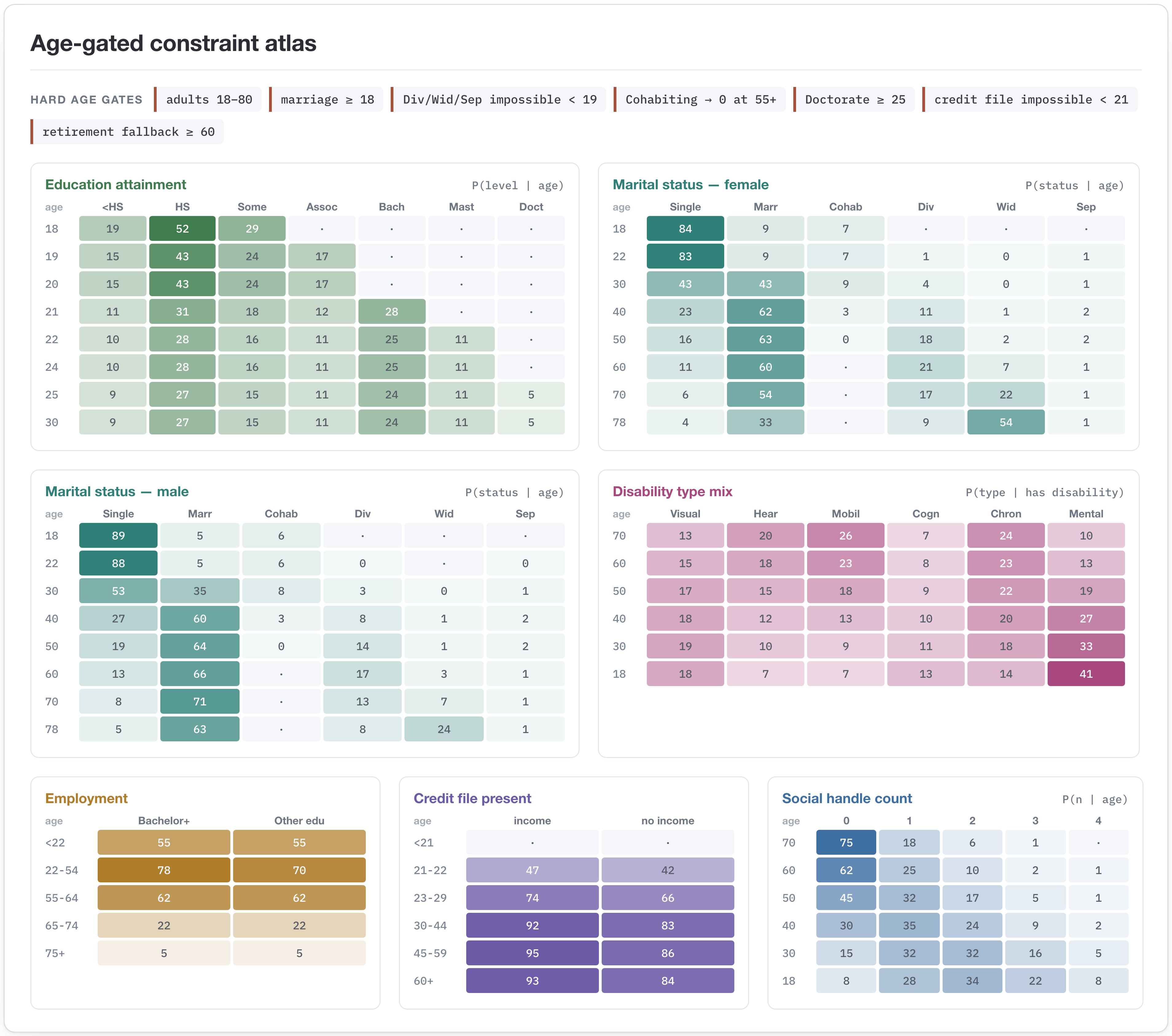}
  \caption{Age-gated constraint atlas for the en-US generator rules. Age is the master gate: each panel is a conditional distribution computed after age gating and renormalization, with cells shown as implemented probabilities multiplied by $100$ and dots marking structural zeros. Higher degrees and former-partner states only become reachable at the configured minimum ages.}
  \label{fig:age_atlas}
\end{figure*}

\begin{figure*}[!htbp]
  \centering
  \includegraphics[width=0.90\textwidth]{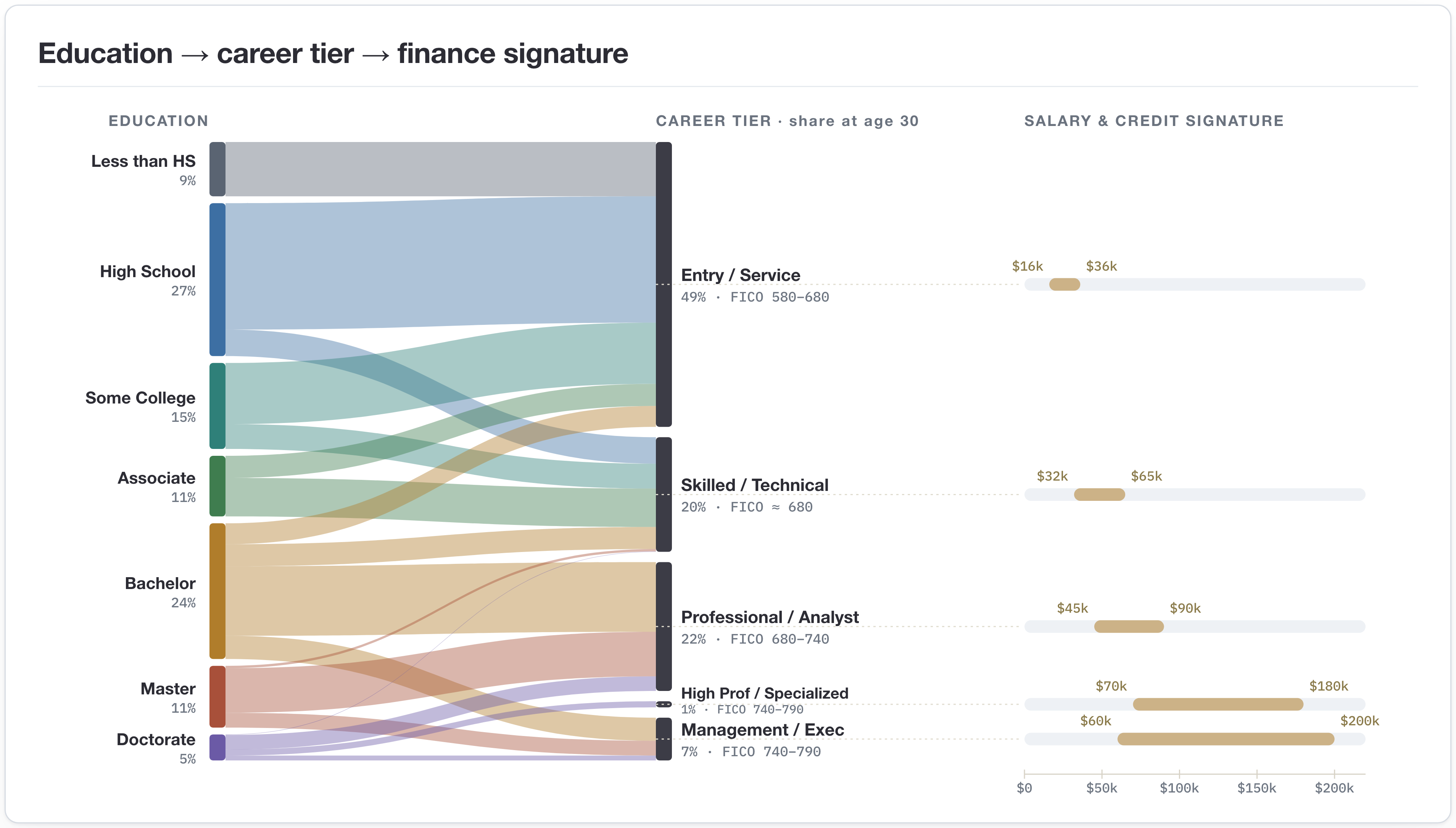}
  \caption{Education to career tier to finance signature. Education indices gate which career tiers a title can be drawn from (ribbons, weighted at age $30$); the chosen tier then bounds salary, which in turn drives the finance tier and credit-score scale. Rare cross-tier crossings are retained as weighted outliers rather than removed.}
  \label{fig:education_career}
\end{figure*}

\begin{figure*}[!htbp]
  \centering
  \includegraphics[width=0.94\textwidth]{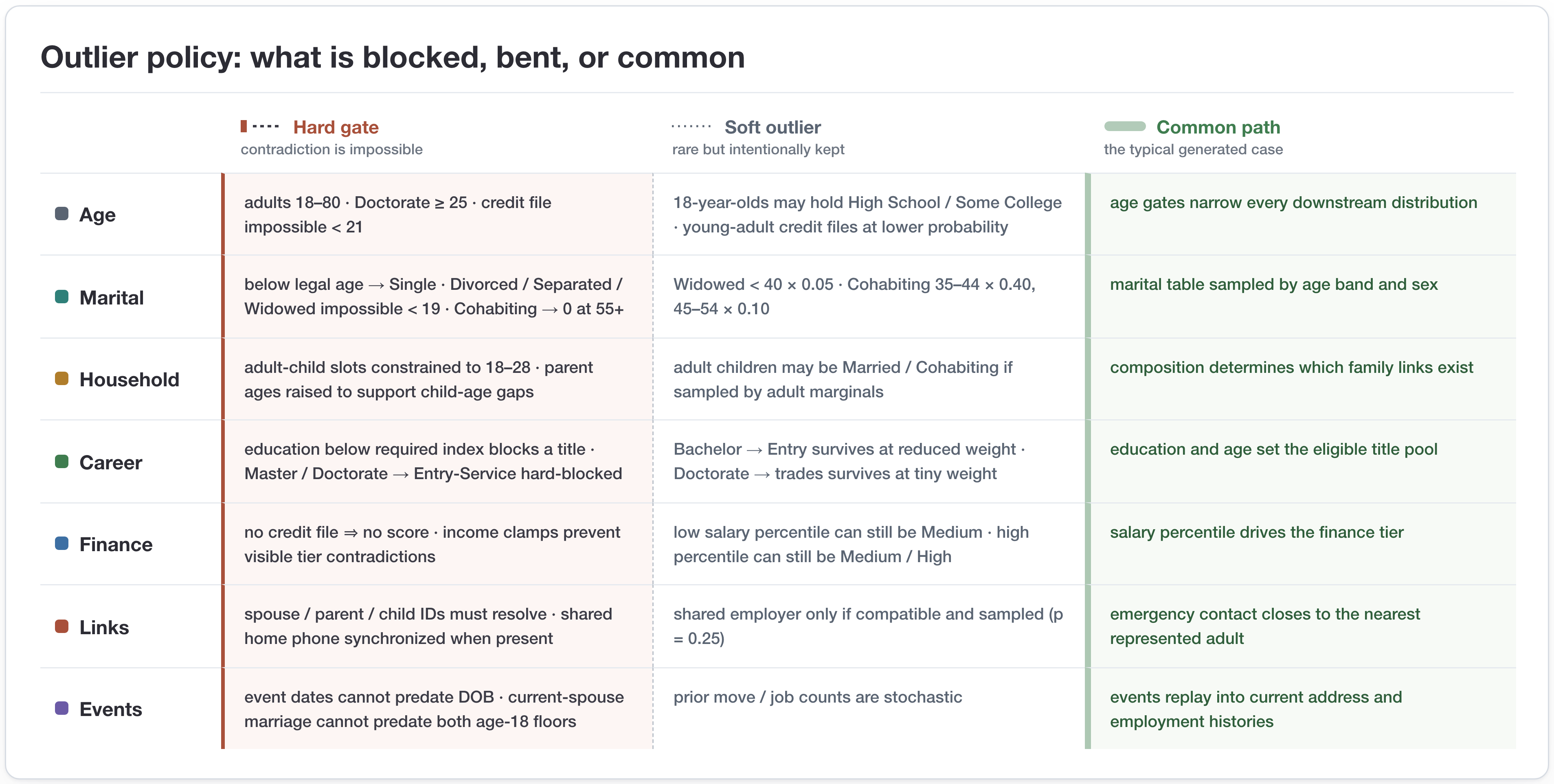}
  \caption{Outlier policy: what is blocked, bent, or common. \pf{} separates the impossible from the merely rare: hard gates remove contradictions outright, weighted bends keep low-probability but plausible cases (for example an $18$-year-old graduate), and ordinary mass covers common combinations.}
  \label{fig:outlier_policy_designer}
\end{figure*}

\paragraph{Generator mechanics.} The generator deliberately preserves some rare combinations after feasibility checks. These weighted outliers are different from contradictions: a doctorate holder may still appear outside high-professional work, a high-income person may still have a lower credit tier, and household members may have different employers unless a shared-employer draw fires. The validator blocks impossible or unsupported combinations, but it does not collapse every profile toward the modal path.

\begin{figure*}[!htbp]
  \centering
  \includegraphics[width=0.60\textwidth]{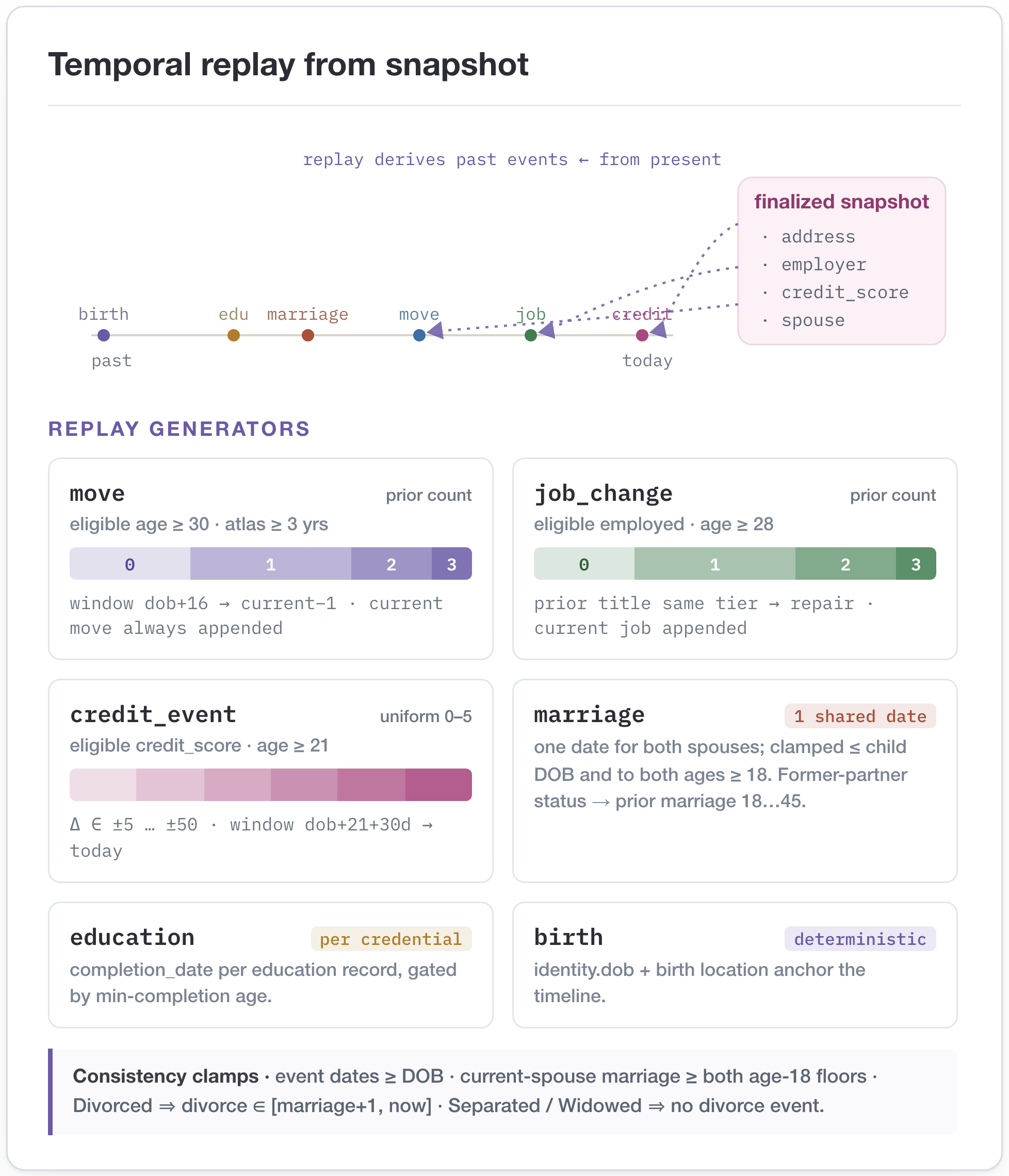}
  \caption{Snapshot-aligned temporal backfill. Histories are reconstructed backward from the finalized snapshot rather than sampled independently; the current address and job are always appended so present state stays consistent with its past. The validator confirms that the latest covered move and job-change events agree with current address and employment; this is partial replay over declared fields, not complete event sourcing.}
  \label{fig:temporal_replay_designer}
\end{figure*}

\paragraph{Assumption ledger.} Table~\ref{tab:assumption-ledger} records the high-level modeling assumptions that downstream users should disclose when using generator outputs.

\begin{table*}[!htbp]
\centering
\small
\begin{tabularx}{\textwidth}{@{}P{0.20\textwidth}P{0.22\textwidth}Y Y@{}}
\toprule
Assumption family & Implementation role & Main risk & Recommended downstream disclosure \\
\midrule
Binary sex/gender conditioning & Conditions selected names and demographic tables & Excludes non-binary identities and can conflate sex and gender & State which field is used; do not infer identity categories beyond the schema \\
Household and surname rules & Produces represented family context & Locale-specific family forms may be underrepresented & Report household types used and avoid generalizing to official demographics \\
Education, occupation, and salary links & Produces internally compatible work and finance fields & Can encode socioeconomic stereotypes & Audit group-conditioned outputs and treat salaries as synthetic plausibility fields \\
Credit and finance tiers & Produces coarse correlated financial state & Not calibrated to real bureau or wealth distributions & Do not use for consequential scoring or population claims \\
Health and disability fields & Adds sparse personal-state attributes & Simplification and stigmatizing associations & Use only when task-relevant; test for harmful correlations \\
Social-platform rules & Adds age- and locale-gated handles & Rapidly changing policies and usage patterns & Record generator version/date and avoid behavioral prevalence claims \\
Names, phones, IDs, addresses & Produces contact-like surface forms & Accidental resemblance or unsafe reuse & Preserve synthetic labels, reserved domains, and collision reports; do not contact or authenticate \\
\bottomrule
\end{tabularx}
\caption{High-level modeling assumptions and disclosure guidance. These are heuristics or priors, not verified descriptions of every locale.}
\label{tab:assumption-ledger}
\end{table*}

\FloatBarrier
\subsection{Released Artifacts: Package and 100K Reference Set}
\label{app:release}

The release ships as both an executable Python package and a fixed 100K reference bundle. Table~\ref{tab:cli} lists the package command surface; the remaining figures and tables describe the reference set, which is an object graph rather than a flat profile table: one canonical Person Object fans out into normalized address, employment, education, event, household, employer, and relationship views.

\begin{table*}[!htbp]
\centering
\small
\begin{tabularx}{\textwidth}{@{}P{0.26\textwidth}Y@{}}
\toprule
Command & Role \\
\midrule
\code{profilefoundry verify} & Generates one deterministic profile per supported locale as a smoke test \\
\code{profilefoundry person} & Generates one inspectable Person Object \\
\code{profilefoundry household} & Generates one linked household \\
\code{profilefoundry scale-households} & Generates household-preserving JSONL at larger scale \\
\code{profilefoundry validate} & Runs configured marginal, consistency, replay, collision, Wikidata, and email checks \\
\code{profilefoundry export} & Writes JSONL, normalized Parquet, manifest, and validation/leakage summaries \\
\bottomrule
\end{tabularx}
\caption{Package command surface.}
\label{tab:cli}
\end{table*}

\begin{figure*}[!htbp]
  \centering
  \includegraphics[width=\textwidth]{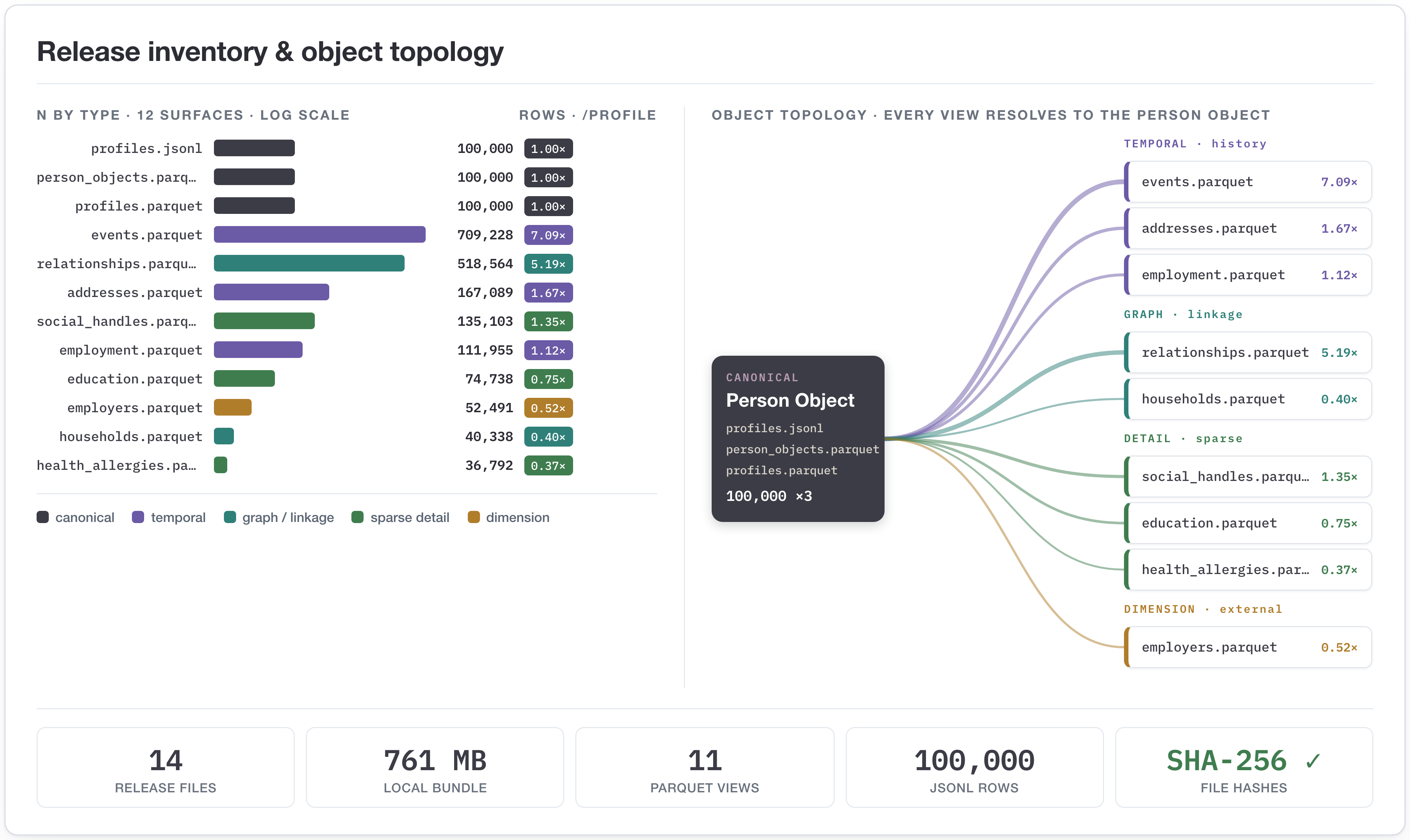}
  \caption{Release inventory and object topology. Raw row counts orient the reader on a log scale spanning $36.8$K--$709$K rows across the twelve normalized views, while the rows-per-profile multipliers and hub edges show that the release is an object graph: one canonical profile expands into addresses, employment, education, typed events, and relationship edges. This panel consolidates what were previously separate inventory and overview figures.}
  \label{fig:release_topology}
\end{figure*}

\begin{table*}[!htbp]
\centering
\small
\begin{tabularx}{\textwidth}{@{}l r r Y@{}}
\toprule
File & Rows & Rows/profile & Purpose \\
\midrule
\code{profiles.jsonl} & 100{,}000 & 1.00 & Canonical nested Person Objects, one JSON object per line. \\
\code{person_objects.parquet} & 100{,}000 & 1.00 & Complete one-row-per-person viewer table, with nested sections encoded as JSON strings. \\
\code{profiles.parquet} & 100{,}000 & 1.00 & Flat scalar snapshot for quick tabular analysis. \\
\code{addresses.parquet} & 167{,}089 & 1.67 & Current and historical addresses with validity intervals and source event IDs. \\
\code{employment.parquet} & 111{,}955 & 1.12 & Employment records, including current and historical jobs. \\
\code{education.parquet} & 74{,}738 & 0.75 & Education records. \\
\code{social_handles.parquet} & 135{,}103 & 1.35 & One row per emitted social handle. \\
\code{health_allergies.parquet} & 36{,}792 & 0.37 & One row per allergy entry. \\
\code{events.parquet} & 709{,}228 & 7.09 & Long-format typed event timeline with payload columns. \\
\code{households.parquet} & 40{,}338 & 0.40 & Household dimension table with members, inferred composition, and shared address. \\
\code{employers.parquet} & 52{,}491 & 0.52 & Employer dimension table aggregated from current and historical employment records. \\
\code{relationships.parquet} & 518{,}564 & 5.19 & Directed relationship graph: family, cohabitation, household membership, and colleague edges. \\
\bottomrule
\end{tabularx}
\caption{Row-counted release views. \code{MANIFEST.json} records file hashes and row counts; the dataset card defines corresponding viewer configs.}
\label{tab:storage_appendix}
\end{table*}

\begin{figure*}[!htbp]
  \centering
  \includegraphics[width=0.98\textwidth]{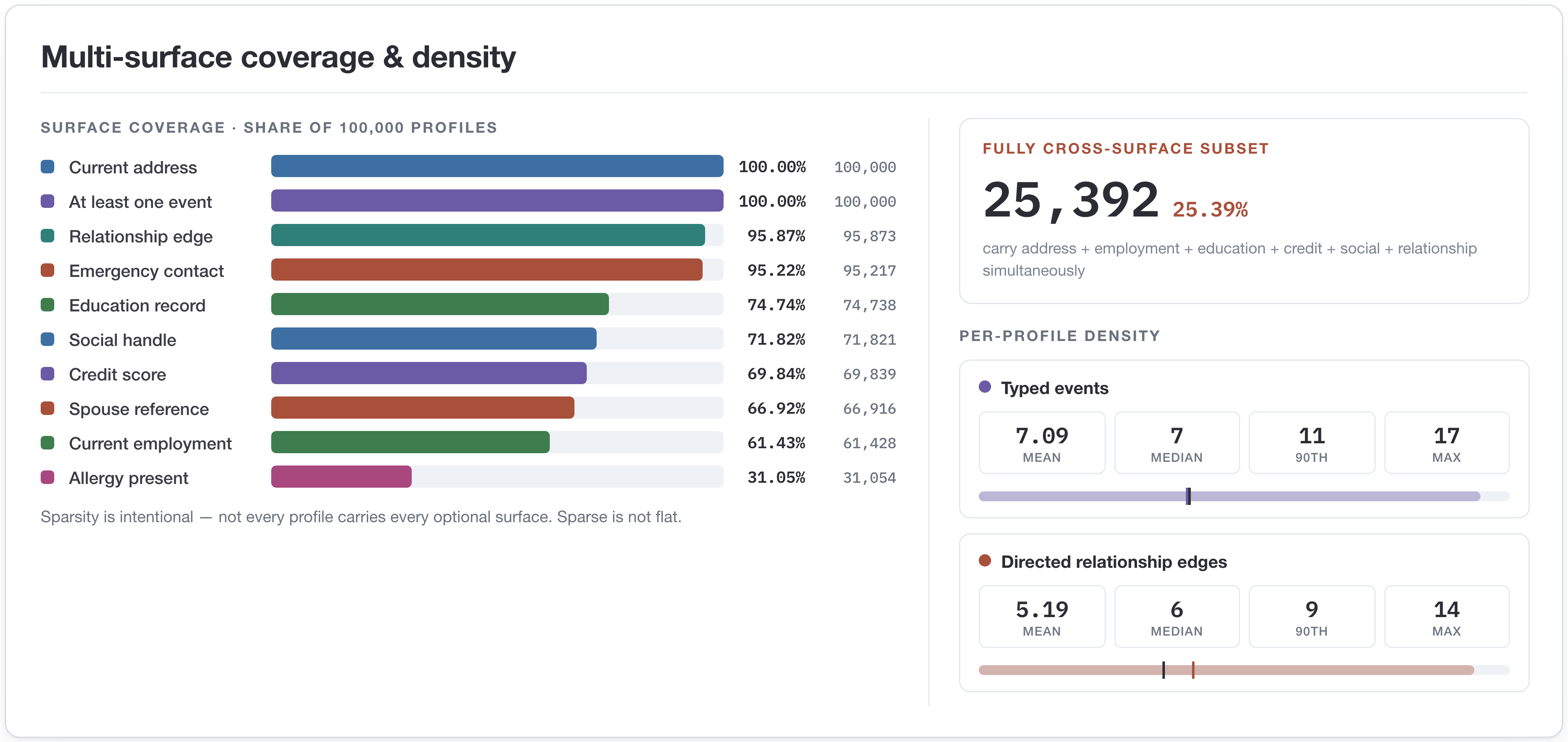}
  \caption{Multi-surface coverage and per-profile density. Every profile carries a current address and event history; $95.9\%$ sit on a relationship edge, and many simultaneously carry employment, education, credit, social, and allergy context. Optional sparsity is explicit, and $25{,}392$ profiles jointly include all six analytical surfaces.}
  \label{fig:profile_coverage}
\end{figure*}

\begin{figure*}[!htbp]
  \centering
  \includegraphics[width=0.98\textwidth]{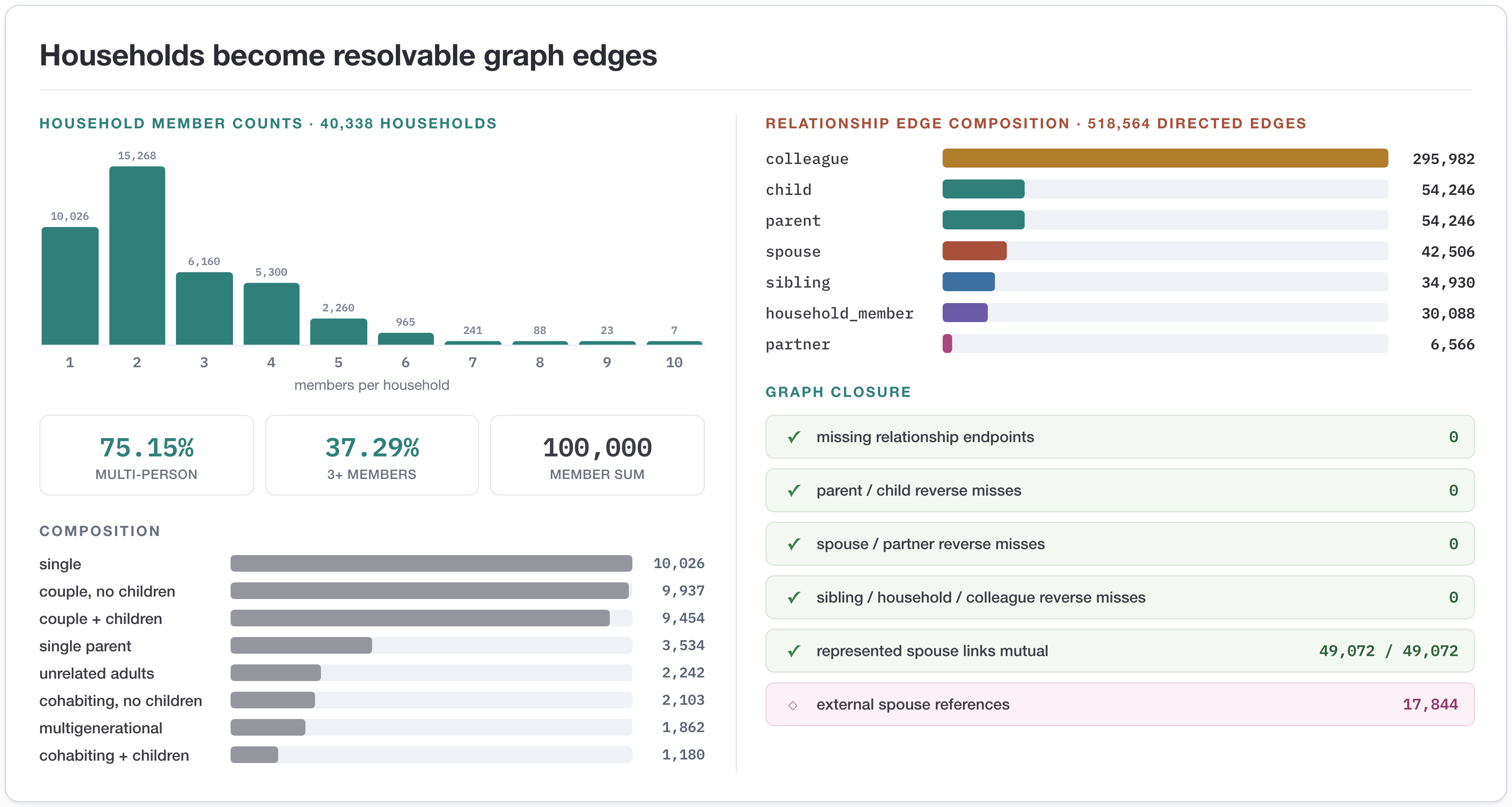}
  \caption{Households resolved into a directed relationship graph. Three quarters of the $40{,}338$ households hold multiple represented adults, and household-membership, family, partner, and colleague relations resolve into $518{,}564$ directed edges. Endpoint and reciprocal-link checks confirm the graph closes; the composition and edge mix are tabulated in Table~\ref{tab:household_graph}.}
  \label{fig:household_graph_fig}
\end{figure*}

\begin{table*}[!htbp]
\centering
\small
\begin{tabularx}{\textwidth}{@{}P{0.23\textwidth}r r P{0.22\textwidth}r r@{}}
\toprule
\multicolumn{3}{c}{\textbf{Households}} & \multicolumn{3}{c}{\textbf{Directed relationships}} \\
\cmidrule(lr){1-3}\cmidrule(lr){4-6}
Composition & Households & Share & Kind & Edges & Share \\
\midrule
single & 10{,}026 & 24.85\% & colleague & 295{,}982 & 57.08\% \\
couple, no children & 9{,}937 & 24.63\% & child & 54{,}246 & 10.46\% \\
couple, adult children & 9{,}454 & 23.44\% & parent & 54{,}246 & 10.46\% \\
single parent, adult children & 3{,}534 & 8.76\% & spouse & 42{,}506 & 8.20\% \\
unrelated adults & 2{,}242 & 5.56\% & sibling & 34{,}930 & 6.74\% \\
cohabiting, no children & 2{,}103 & 5.21\% & household member & 30{,}088 & 5.80\% \\
multigenerational & 1{,}862 & 4.62\% & partner & 6{,}566 & 1.27\% \\
cohabiting, adult children & 1{,}180 & 2.93\% & \textbf{Total} & \textbf{518{,}564} & \textbf{100\%} \\
\bottomrule
\end{tabularx}
\caption{Household compositions and relationship-edge mix. Of 40{,}338 households, 30{,}312 are multi-person and 15{,}044 contain at least three represented adults.}
\label{tab:household_graph}
\end{table*}

\begin{figure*}[!htbp]
  \centering
  \includegraphics[width=0.98\textwidth]{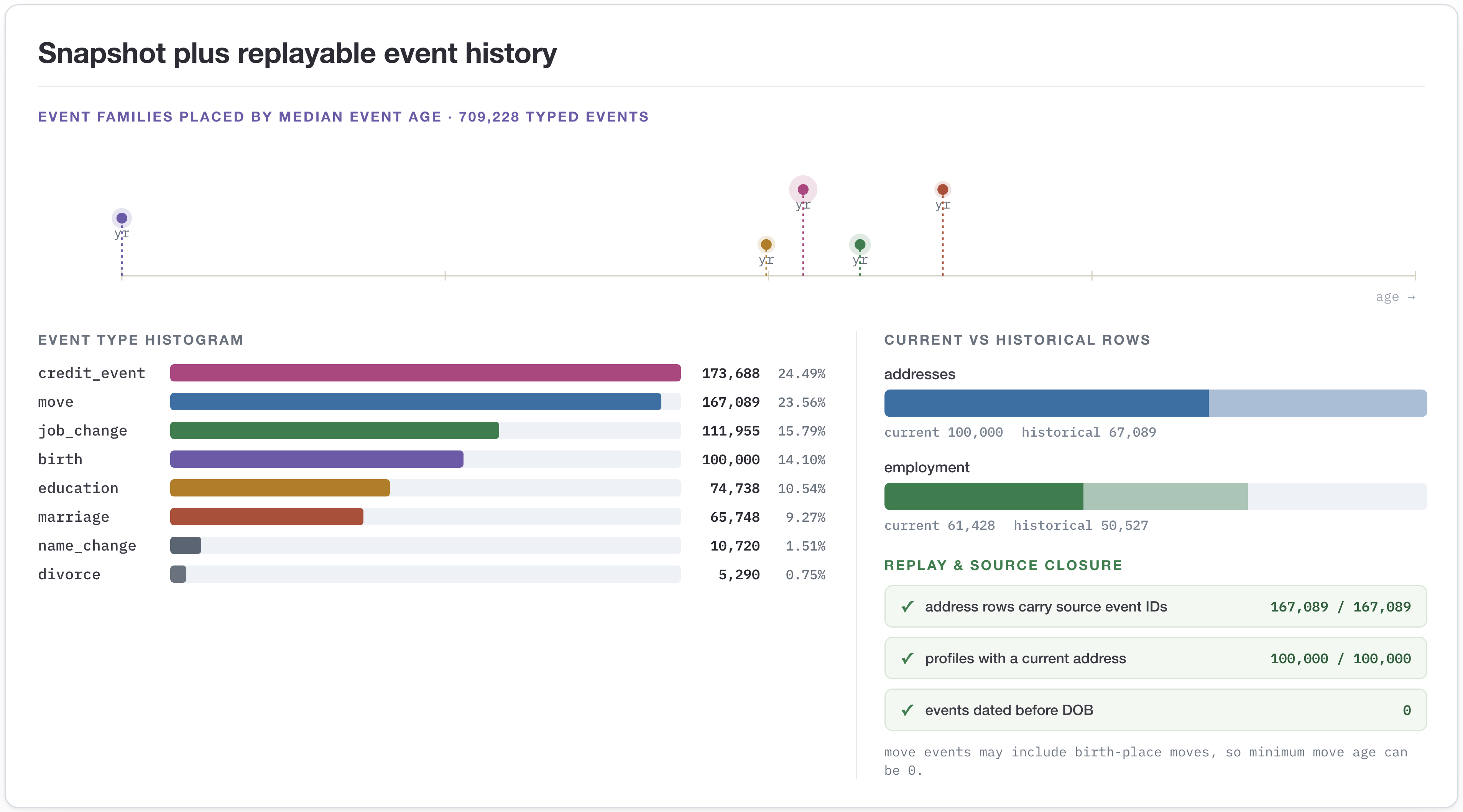}
  \caption{Temporal release surface. Typed events project into address and employment histories: every address row carries source-event identifiers, every profile has exactly one current address, and no event predates date of birth. The typed-event composition and selected age summaries are tabulated in Table~\ref{tab:event_inventory}.}
  \label{fig:temporal_surface}
\end{figure*}

\begin{table*}[!htbp]
\centering
\small
\begin{tabularx}{\textwidth}{@{}P{0.15\textwidth} r r P{0.18\textwidth} r Y@{}}
\toprule
\multicolumn{3}{c}{\textbf{Event inventory}} & \multicolumn{3}{c}{\textbf{Temporal age summaries}} \\
\cmidrule(lr){1-3}\cmidrule(lr){4-6}
Event type & Events & Share & Check & Age & Interpretation \\
\midrule
\code{credit_event} & 173{,}688 & 24.49\% & birth median & 0.00 & anchored to DOB \\
\code{move} & 167{,}089 & 23.56\% & education median & 19.93 & completion or event timing \\
\code{job_change} & 111{,}955 & 15.79\% & job-change median & 22.83 & employment timing \\
\code{birth} & 100{,}000 & 14.10\% & marriage median & 25.39 & represented commitment \\
\code{education} & 74{,}738 & 10.54\% & credit-event minimum & 21.07 & configured age gate \\
\code{marriage} & 65{,}748 & 9.27\% & move minimum & 0.00 & can encode birthplace/current-address origin \\
\code{name_change} & 10{,}720 & 1.51\% & & & \\
\code{divorce} & 5{,}290 & 0.75\% & & & \\
\bottomrule
\end{tabularx}
\caption{Typed-event composition and selected temporal sanity summaries.}
\label{tab:event_inventory}
\end{table*}

\begin{figure*}[!htbp]
  \centering
  \includegraphics[width=0.98\textwidth]{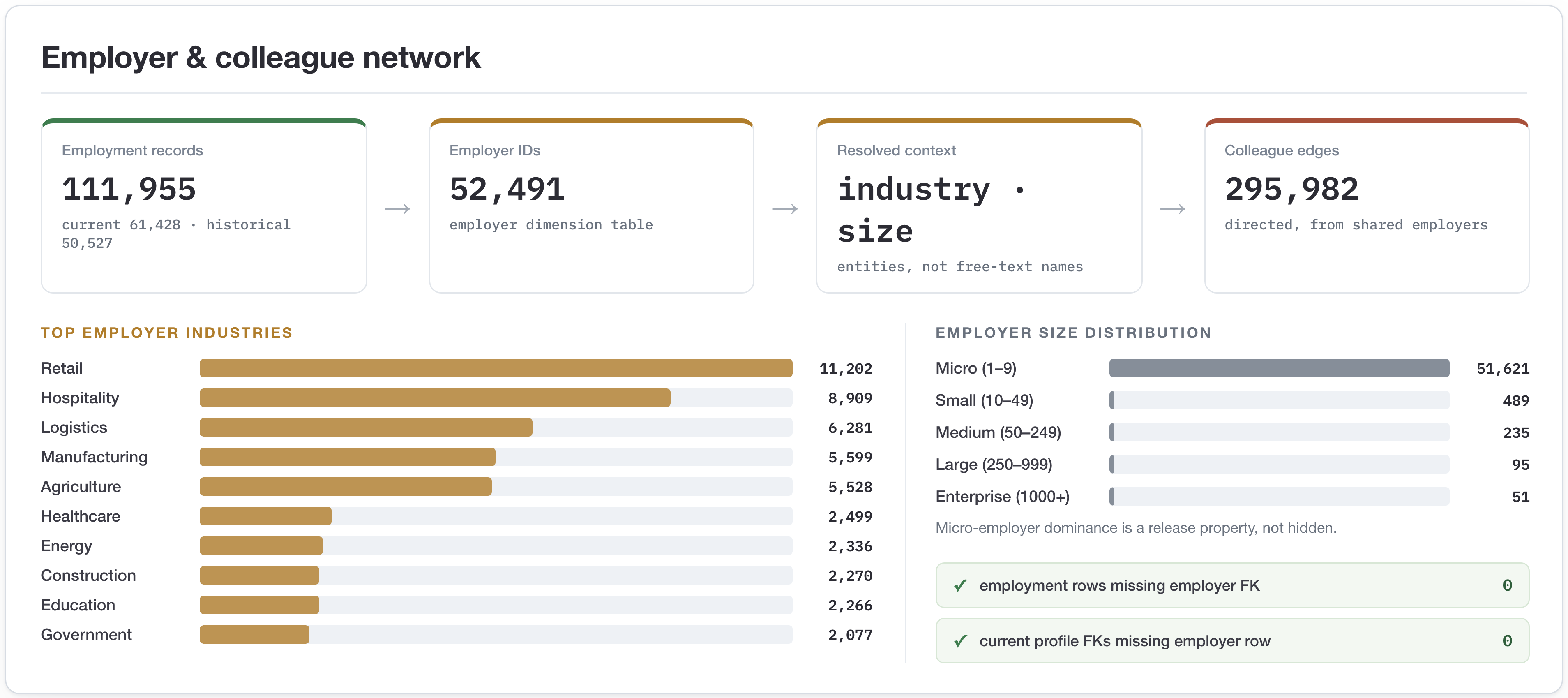}
  \caption{Employer context exported as resolvable entities rather than free-text names. Current and historical employment resolves to $52{,}491$ employer IDs and produces $295{,}982$ directed colleague edges with zero missing foreign keys, so shared-employer co-membership can be queried as graph structure rather than inferred from matching strings.}
  \label{fig:employer_network}
\end{figure*}

\begin{table*}[!htbp]
\centering
\small
\begin{tabularx}{\textwidth}{@{}l Y Y@{}}
\toprule
Schema group & Representative fields & Release evidence \\
\midrule
Identity & Names, DOB, gender, marital status, nationality, birth city/country & Pydantic types; leakage audits over name-DOB and name-city tuples; display-name collision checks. \\
Contact & Personal/work emails, phones, social handles & Reserved domains, syntax audit, uniqueness counts, platform age/banned-platform checks. \\
Addresses & Line, city, region, postcode, country, valid-from/to & One current address maximum; current and historical rows projected from move events. \\
Employment & Employer ID/name, title, salary, valid-from/to & Employer foreign keys, current/historical rows, current employee counts, colleague edges. \\
Education & Institution, degree, field, start/completion dates & Education events, field-of-study checks, age-at-completion gates, and marginal validation. \\
Finance/health/IDs & Income, net worth, credit score/scale, allergies, emergency contact, government IDs & Credit scale validators; age gates; allergy export; uniqueness for issued IDs. \\
Linkage & Household ID, spouse/partner, parents, children, siblings & Household/member rows, family edges, directed relationship table. \\
Timeline/provenance & \code{events[]}, \code{documents[]}, generation metadata & Replay checks, UUID5 event IDs, seed/version/date/manifest hash. \\
\bottomrule
\end{tabularx}
\caption{Schema coverage and corresponding evidence.}
\label{tab:schema_coverage}
\end{table*}

\FloatBarrier
\paragraph{Release use guidance.} Table~\ref{tab:uses_appendix} summarizes recommended, caveated, and discouraged uses of the released package and reference population.

\begin{table*}[!htbp]
\centering
\small
\begin{tabularx}{\textwidth}{@{}l Y@{}}
\toprule
Category & Guidance \\
\midrule
Recommended uses & Agent state tracking, CRM/KYC test harnesses, document extraction prototypes, record-linkage experiments, synthetic identity handling in privacy research, dataset-card and validation-method case studies. \\
Allowed with caveats & Email, form, and document workflows that require public email-domain deliverability or rendered artifacts; users must substitute domains or create documents and then run separate audits. \\
Not recommended & Training consequential decision systems about real people, simulating official population statistics, impersonation, credential testing, or treating the release as privacy-proof data. \\
\bottomrule
\end{tabularx}
\caption{Use guidance for downstream users.}
\label{tab:uses_appendix}
\end{table*}

\FloatBarrier
\subsection{Audit}
\label{app:audit}

\begin{figure*}[!htbp]
  \centering
  \includegraphics[width=0.72\textwidth]{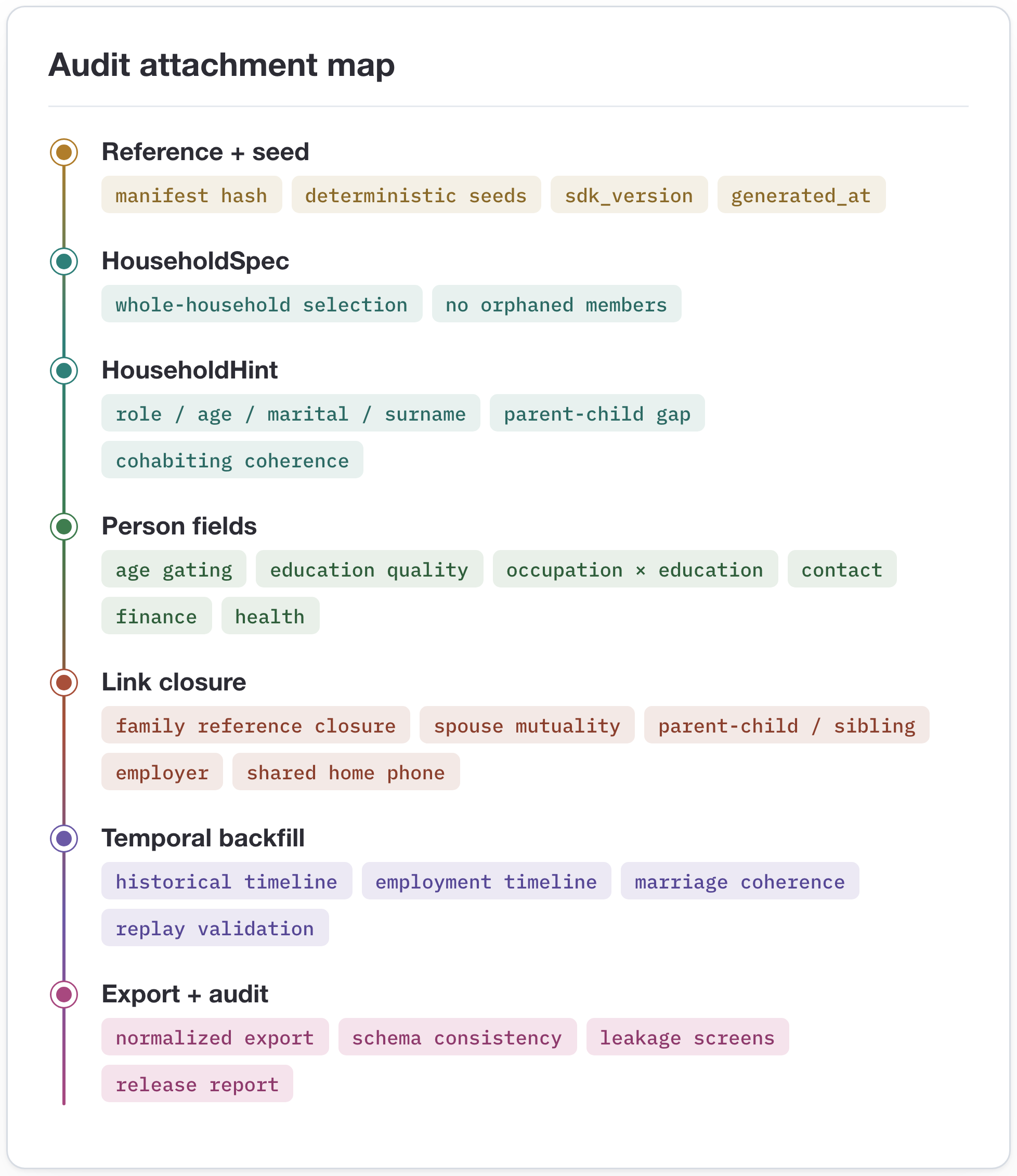}
  \caption{Audit attachment map. Every generation stage carries its own validator family into the release audit, so validation, replay, referential-integrity, manifest, and leakage checks attach to the part of the release object they verify rather than to a single flat table.}
  \label{fig:audit_attachment_designer}
\end{figure*}

Audit evidence is reported as several distinct forms rather than a single pass/fail summary. Each generation stage carries its own validator family into the release audit (Figure~\ref{fig:audit_attachment_designer}): distributional fit against public marginals, referential and temporal closure, leakage and collision screening, structural invariants, and reproducibility are stated separately.

\begin{figure*}[!htbp]
  \centering
  \includegraphics[width=0.98\textwidth]{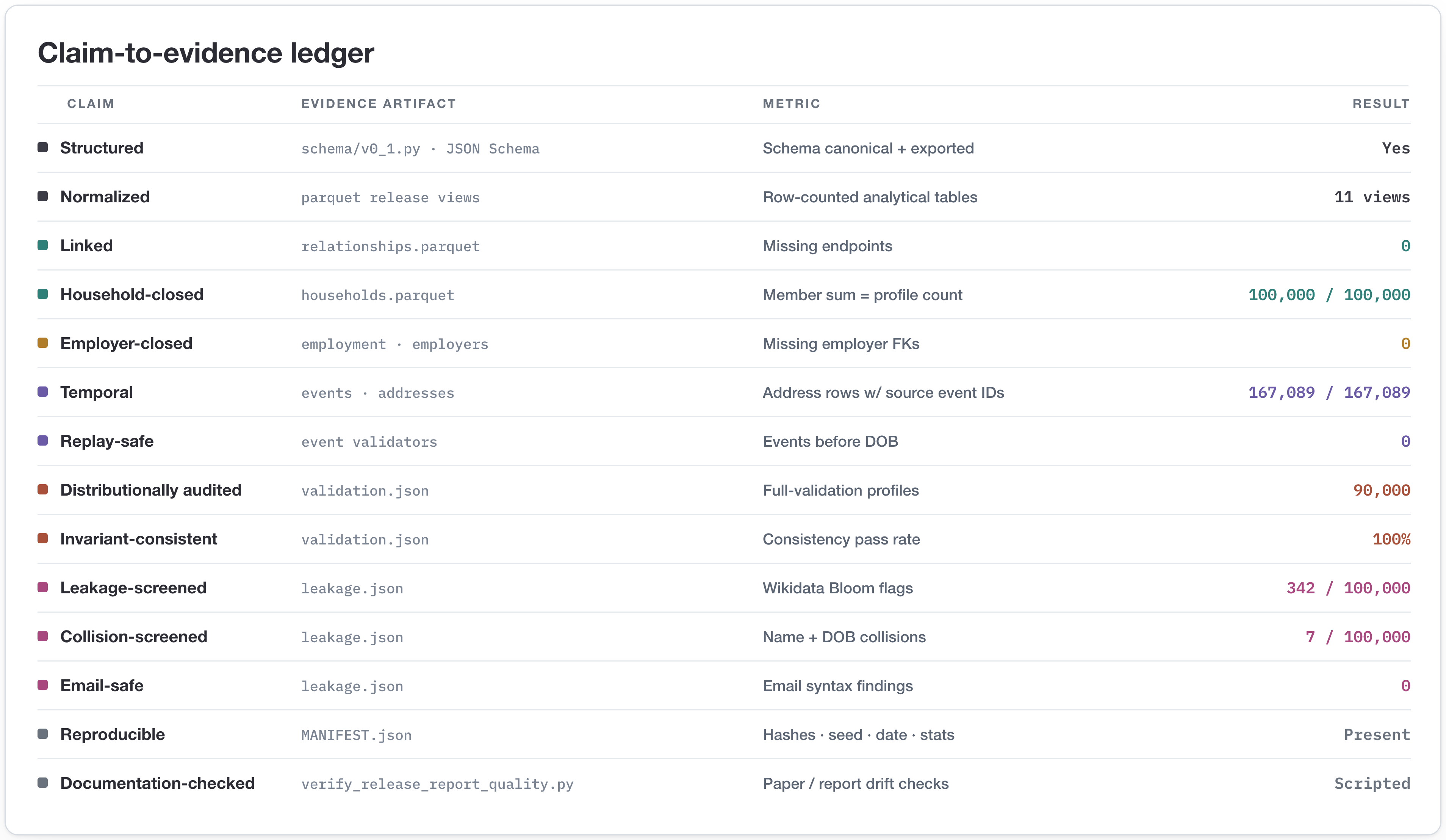}
  \caption{Claim-to-evidence ledger. Each headline capability of the release (structured, executable, consistent, linked, temporal, leakage-audited, reproducible, and documentation-checked) maps to a concrete public artifact and a verifying metric or check. This visual ledger summarizes the same correspondence detailed textually in Table~\ref{tab:artifact_evidence}.}
  \label{fig:claim_evidence_ledger}
\end{figure*}

\paragraph{Evidence overview.} Figure~\ref{fig:claim_evidence_ledger} and Table~\ref{tab:artifact_evidence} map each headline claim about the release to the concrete public artifact that backs it and to the metric or check that verifies it. The figure is a visual ledger of this correspondence; the table states the same mapping in full.

\begin{table*}[!htbp]
\centering
\small
\begin{tabularx}{\textwidth}{@{}l Y Y@{}}
\toprule
Claim & Evidence artifact & Notes \\
\midrule
Structured & Schema source plus exported JSON Schema & Pydantic source of truth plus JSON Schema export. \\
Executable & Python package, CLI, and repository scripts & Command surface for single profiles, households, validation, exports, and full release runs. \\
Declared-suite consistency & Validation report plus invariant tests & 100\% pass in US/UK/IN/CA/AU full-validation locales. \\
Linked & \code{households.parquet}, \code{employers.parquet}, \code{relationships.parquet} & 40{,}338 households, 52{,}491 employer IDs, 518{,}564 directed edges. \\
Temporal & \code{events.parquet}; replay validator & 709{,}228 typed events; latest covered move/job events agree with snapshot. \\
Leakage audited & Release leakage report & Self-collision, Wikidata Bloom, and reserved-domain email evidence. \\
Reproducible & generation metadata, fixture, manifest hashes & Seed/version/date/reference hash identify the release. \\
Documentation-checked & Report-quality verifier & Counts, validation, leakage, figure, and manuscript claims are checked against reports. \\
\bottomrule
\end{tabularx}
\caption{Artifact evidence summary.}
\label{tab:artifact_evidence}
\end{table*}

\begin{figure*}[!htbp]
  \centering
  \includegraphics[width=0.96\textwidth]{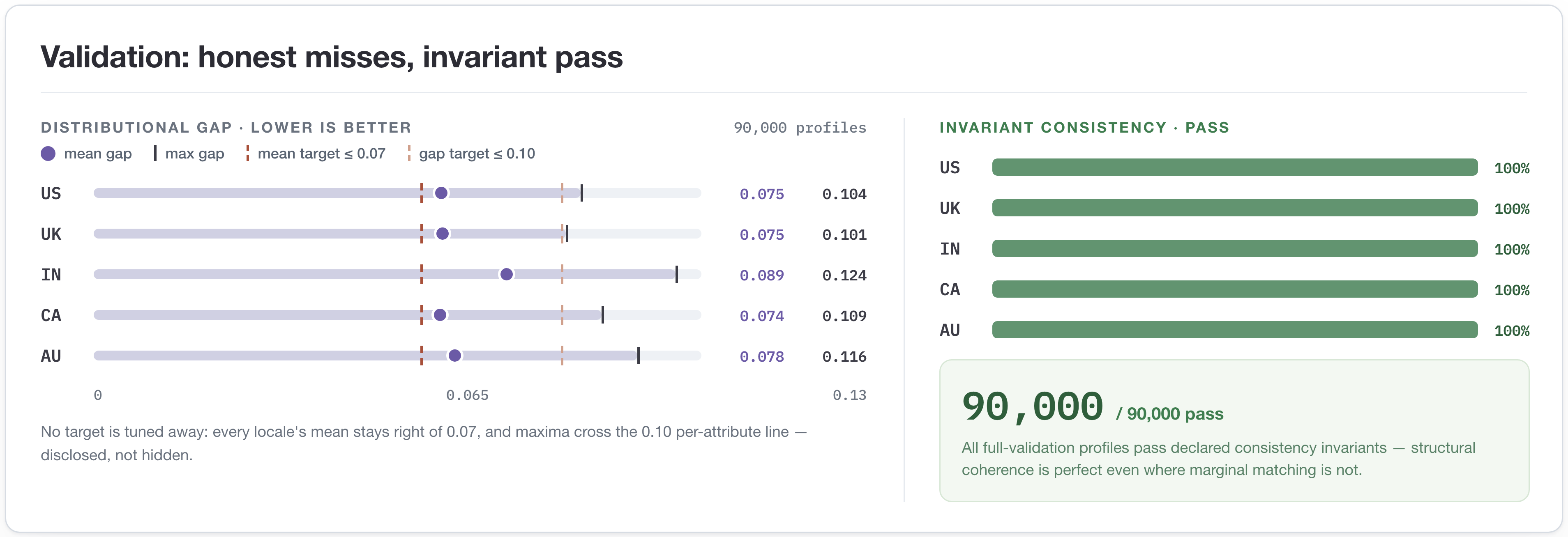}
  \caption{Validation target audit: honest misses alongside invariant pass. Distributional fit is disclosed separately from invariant consistency; all $90{,}000$ full-validation profiles pass the declared-consistency checks, while marginal gaps that exceed the per-attribute target are published rather than tuned away. Per-locale detail is given in Table~\ref{tab:validation_interpretation}.}
  \label{fig:validation_target_bars}
\end{figure*}

\begin{table*}[!htbp]
\centering
\small
\begin{tabularx}{\textwidth}{@{}l r r r r r r Y@{}}
\toprule
Locale & $N$ & age M & age F & education & marital & mean & Interpretation \\
\midrule
US & 35{,}000 & 0.104 & 0.100 & 0.004 & 0.091 & 0.075 & Close to target; age-male slightly above 0.10 after rounding. \\
UK & 20{,}000 & 0.101 & 0.093 & 0.011 & 0.095 & 0.075 & Close to target; age-male slightly above 0.10. \\
IN & 20{,}000 & 0.124 & 0.124 & 0.002 & 0.104 & 0.089 & Largest disclosed mismatch; age and marital marginals need future work. \\
CA & 8{,}000 & 0.109 & 0.102 & 0.007 & 0.080 & 0.074 & Closest mean gap, but age gaps remain above the per-attribute threshold. \\
AU & 7{,}000 & 0.116 & 0.102 & 0.010 & 0.083 & 0.078 & Age gaps exceed target; education and marital gaps are small. \\
\bottomrule
\end{tabularx}
\caption{Validation interpretation by locale. The declared consistency and invariant pass rate is 100\% for all five full-validation locales.}
\label{tab:validation_interpretation}
\end{table*}

The full-validation report uses public age-by-sex, education, and marital-status reference tables for US, UK, IN, CA, and AU. The metric is a maximum absolute bucket-share discrepancy ($L_{\infty}$ marginal gap), not a KS statistic. The report does not tune the generator repeatedly against the target because that would overfit the release to its own audit. IE, NZ, and PH are included in the release but excluded from the locked marginal-fit and consistency table in v1.0.

\begin{table*}[!htbp]
\centering
\small
\begin{tabularx}{\textwidth}{@{}l l Y@{}}
\toprule
Locale & Validation tier & Reference sources recorded in the project manifest \\
\midrule
US & full & ACS 2023 one-year tables, SSA names, US Census surnames, ZCTA and business-pattern references. \\
UK & full & ONS 2021 Census tables, Family Resources Survey/HBAI income references, ONS baby-name and surname references. \\
IN & full & Census of India 2011 tables, NSS/PLFS 2022--23 and related government statistical sources. \\
CA & full & Statistics Canada 2021 Census tables and income/labor-force references. \\
AU & full & ABS 2021 Census and Survey of Income and Housing references. \\
IE & light & CSO Census references, bootstrap marginals only in v1.0. \\
NZ & light & Stats NZ Census references, bootstrap marginals only in v1.0. \\
PH & light & PSA Census references, bootstrap marginals only in v1.0. \\
\bottomrule
\end{tabularx}
\caption{Reference-data provenance summary. The generator can fall back to committed bootstrap marginals; richer derived tables are regenerated from source APIs when available.}
\label{tab:reference_sources}
\end{table*}

\begin{figure*}[!htbp]
  \centering
  \includegraphics[width=0.98\textwidth]{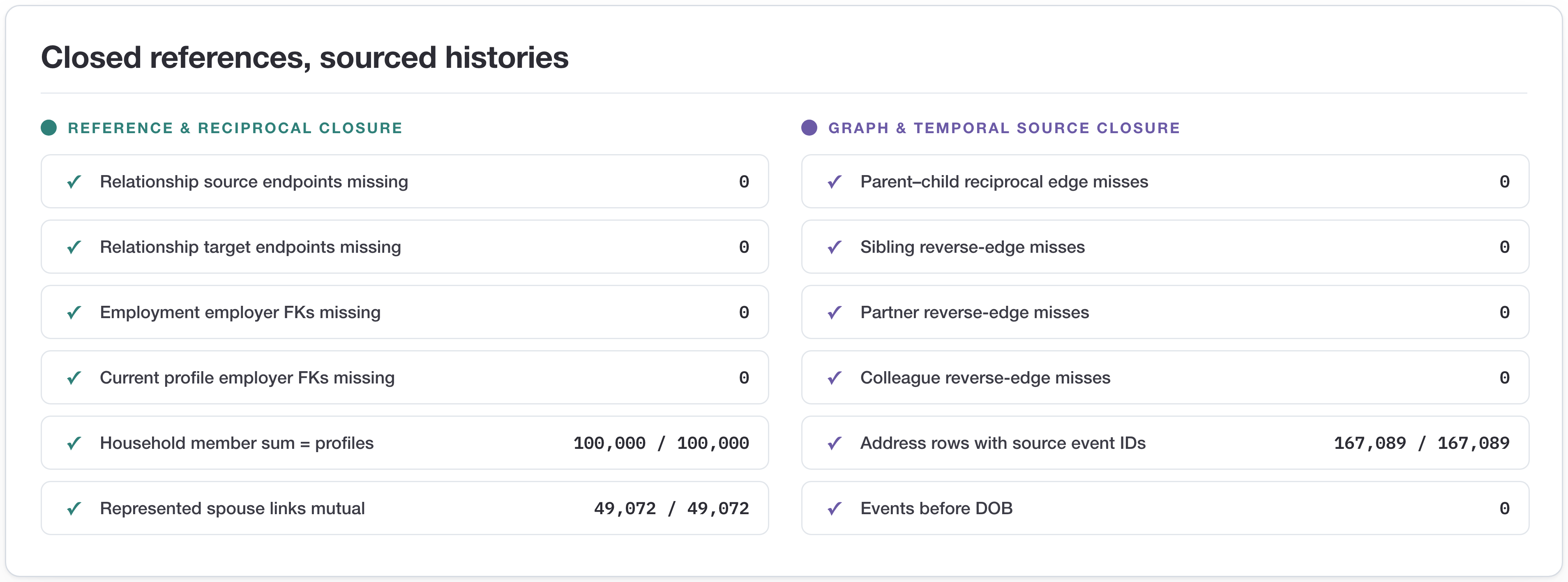}
  \caption{Release-wide referential and temporal closure. Relationship endpoints resolve to released profiles, employer references resolve to employer rows, and represented spouse and parent--child links close reciprocally; exact checks are stated individually rather than summarized as ``no errors.'' The full check inventory is given in Table~\ref{tab:closure_checks}.}
  \label{fig:closure_temporal}
\end{figure*}

\begin{table*}[!htbp]
\centering
\small
\begin{tabularx}{\textwidth}{@{}Y r Y r@{}}
\toprule
\textbf{Graph or entity check} & \textbf{Result} & \textbf{Temporal/source check} & \textbf{Result} \\
\midrule
Missing relationship source endpoints & 0 & Address rows with source event IDs & 167{,}089 / 167{,}089 \\
Missing relationship target endpoints & 0 & Current addresses & 100{,}000 / 100{,}000 \\
Missing employer foreign keys & 0 & Events dated before DOB & 0 \\
Represented spouse links mutual & 49{,}072 / 49{,}072 & Current address rows & 100{,}000 \\
Parent--child reciprocal misses & 0 & Historical address rows & 67{,}089 \\
Sibling reverse-edge misses & 0 & Current employment rows & 61{,}428 \\
Partner reverse-edge misses & 0 & Historical employment rows & 50{,}527 \\
Colleague reverse-edge misses & 0 & Total typed events & 709{,}228 \\
Household member sum & 100{,}000 / 100{,}000 & Events/profile, mean/median/90th/max & 7.09 / 7 / 11 / 17 \\
\bottomrule
\end{tabularx}
\caption{Referential and temporal closure checks over the fixed release. External spouse references (17{,}844) are explicit sentinel cases rather than missing endpoints.}
\label{tab:closure_checks}
\end{table*}

\begin{figure*}[!htbp]
  \centering
  \includegraphics[width=0.98\textwidth]{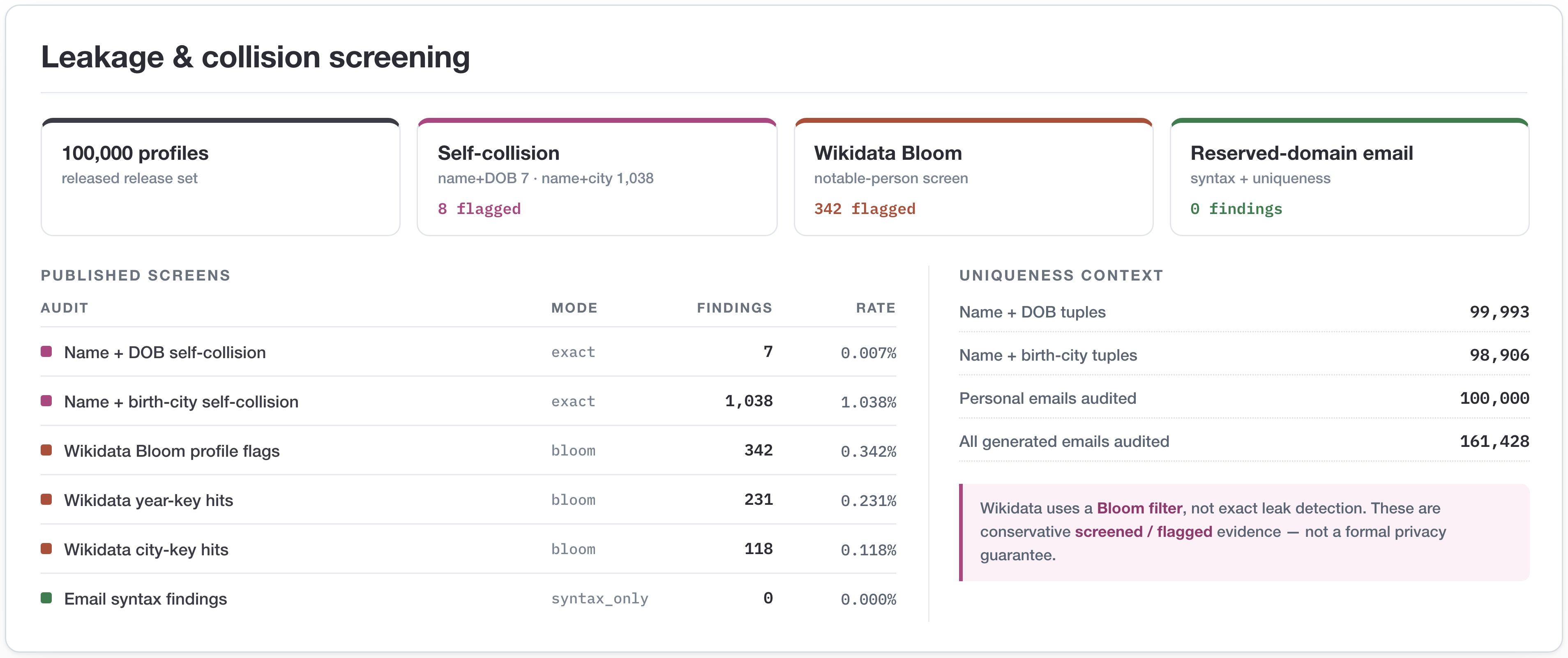}
  \caption{Collision and coincidence screening. The release publishes exact within-release collision checks, a Wikidata Bloom notable-person screen, and reserved-domain email syntax and uniqueness evidence. These are separate screens over common names and places and do not constitute a formal privacy guarantee; denominators and interpretation are given in Table~\ref{tab:leakage_interpretation}.}
  \label{fig:leakage_screening}
\end{figure*}

\begin{table*}[!htbp]
\centering
\small
\begin{tabularx}{\textwidth}{@{}l r r Y@{}}
\toprule
Audit & Findings & Unique tuples & What the result means \\
\midrule
name + DOB & 7 & 99{,}993 & Exact within-release duplicates are rare at 100K scale. \\
name + birth city & 1{,}038 & 98{,}906 & Expected to be higher because city is lower-cardinality than date of birth. \\
personal email & 0 & 100{,}000 & Every personal email local-part/domain pair is unique in the release. \\
Wikidata Bloom scan & 342 & -- & Conservative notable-person coincidence screen, not proof of copying. \\
Email syntax audit & 0 & -- & Release emails use reserved domains and pass syntax checks. \\
\bottomrule
\end{tabularx}
\caption{Interpretation of the release leakage and collision audits.}
\label{tab:leakage_interpretation}
\end{table*}

\paragraph{Leakage methodology.} The Wikidata filter is constructed over humans with a known birth date and at least five sitelinks. The released filter covers birth years 1850--2015, contains 683{,}897 records, and indexes 667{,}179 name-year keys plus 631{,}907 name-city keys. Bloom filters introduce false positives; the configured target false-positive rate is $10^{-4}$. Because the filter is a screening tool over common names and places, its output is best interpreted as a conservative coincidence rate.

The HIBP prototype audit is intentionally not a release metric. The public-email-domain prototype demonstrated that realistic \code{first.last@provider} patterns can overlap breached-account corpora even when no real identity was copied. The current release changes the design by using reserved \code{*.profilefoundry.example} domains, so the correct release audit is syntax and uniqueness rather than breached-account lookup.

\begin{table*}[!htbp]
\centering
\small
\begin{tabularx}{\textwidth}{@{}l Y@{}}
\toprule
Invariant family & Examples \\
\midrule
Pydantic structural checks & No extra fields; event dates sorted; events do not predate DOB; at most one current address and at most one current employment. \\
Age gates & Minors cannot have employment, work emails, credit scores, or finance status; under-13 profiles cannot have social handles. The release is adult-only, so these are regression checks. \\
Locale/contact checks & US territories excluded from US state field; phone numbers normalized; mobile prefixes checked by locale; TikTok blocked for IN profiles; reserved email domains. \\
Partial replay checks & Birth event matches DOB; education events match records; latest job-change matches current employment; latest move matches current address; represented married profiles have marriage events. \\
Referential integrity & Employment records resolve to employer rows; relationship endpoints resolve to profiles; household member lists resolve to profiles; whole-household selection prevents orphaned family references. \\
Collision guards & Personal emails are unique; issued synthetic national IDs, passports, and licenses are collision-checked; same-household display-name ambiguity is guarded. \\
Reproducibility and drift & Pinned fixture byte-compares under deterministic seed/version/manifest hash; release-report verifier checks counts, validation, leakage, figures, paper, and docs. \\
\bottomrule
\end{tabularx}
\caption{Invariant families used to support the declared-suite consistency claim.}
\label{tab:invariants}
\end{table*}

\begin{figure*}[!htbp]
  \centering
  \includegraphics[width=0.72\textwidth]{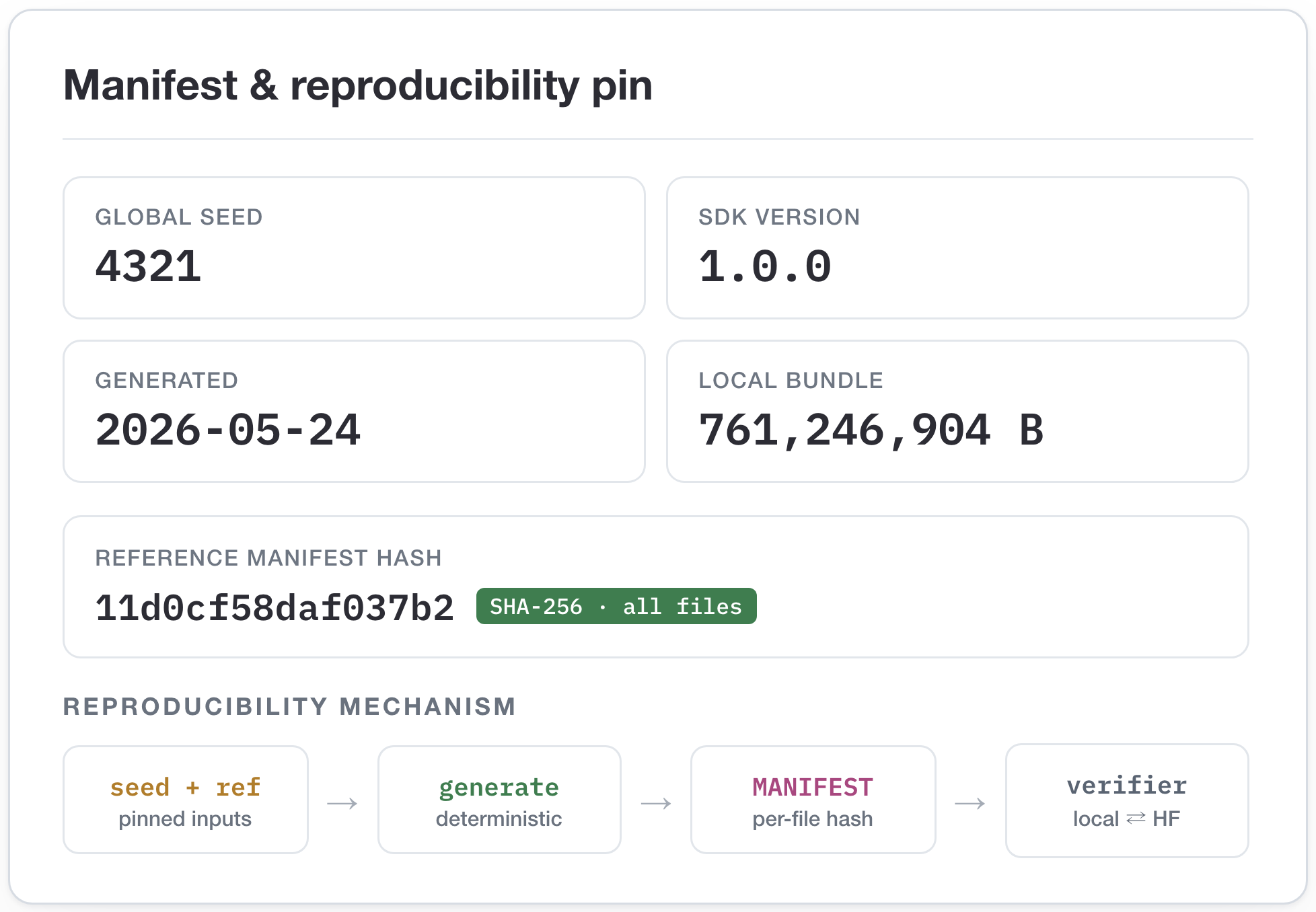}
  \caption{Reproducibility pin. The release records the global seed, generation date, export timestamp, reference-data hash, row counts, and per-file SHA-256 hashes; a release verifier compares the local and published bundles so that any drift is caught. The corresponding command-level checklist is given in Table~\ref{tab:repro_checklist}.}
  \label{fig:reproducibility}
\end{figure*}

\begin{table*}[!htbp]
\centering
\small
\begin{tabularx}{\textwidth}{@{}l Y@{}}
\toprule
Check & Evidence or command \\
\midrule
Install package & \code{pip install profilefoundry}; source installs use \code{pip install -e .}. \\
Smoke CLI & \code{profilefoundry verify}; generates one profile per supported locale. \\
Generate linked household & \code{profilefoundry household --locale UK --seed 4321 --seq 7}. \\
Regenerate release bundle & \code{python scripts/run_full_core.py --generation-date 2026-05-24 --exported-at 2026-05-24 --skip-hibp}. \\
Generate normalized subset & \code{profilefoundry export --out /tmp/pf_core --n-per-locale 1000 --generation-date 2026-05-24 --skip-hibp}. \\
Verify deterministic fixture & \code{python scripts/verify_reproducibility.py}; committed fixture guards against byte-level drift. \\
Verify release reports & \code{python scripts/verify_release_report_quality.py}; catches stale counts, validation, leakage, manuscript, and figure claims. \\
Verify release bundle & \code{python scripts/verify_hf_release_current.py}; compares manifest, row counts, hashes, and referential integrity. \\
Rebuild schema & \code{python scripts/export_schema.py}; emits \code{schemas/person_v0_1.schema.json}. \\
Rebuild Wikidata Bloom & \code{python scripts/ingest_wikidata_persons.py --scope sample}; long-running due to public endpoint rate limits. \\
Run tests & \code{pytest -x -q}; covers schema, factory, linkage, validation, normalized export, leakage, and reproducibility tests. \\
\bottomrule
\end{tabularx}
\caption{Reproducibility checklist. The fixed data release is keyed by seed, generation date, exported timestamp, manifest identifier, and per-file hashes; external version labels must be reconciled before archival submission.}
\label{tab:repro_checklist}
\end{table*}

\FloatBarrier
\subsection{Comparison with Adjacent Resources}
\label{app:related}

\paragraph{Comparison rubric.} The main-text comparison is intentionally descriptive. The rubric below distinguishes public artifacts from generation-time internals, persistent identifiers from ordinary co-occurrence, explicit state change from timestamps, localized fields from multilingual text, and formal privacy from quality or provenance checks. The resource rows are kept individual rather than grouped into heterogeneous families.

\begin{table*}[!htbp]
\centering
\scriptsize
\setlength{\tabcolsep}{3.1pt}
\renewcommand{\arraystretch}{0.94}
\begin{tabularx}{\textwidth}{@{}P{0.115\textwidth}P{0.235\textwidth}P{0.205\textwidth}P{0.205\textwidth}Y@{}}
\toprule
\textbf{Resource} & \textbf{Publicly exposed artifact} & \textbf{Persistent linking} & \textbf{Temporal mechanism} & \textbf{Primary scope} \\
\midrule
This work & Typed Person Objects in JSONL and keyed Parquet views & Household, family, employer, and colleague identifiers and edges & Snapshot-aligned typed events with partial replay and source checks & Domain-general personal-state source layer \\
Pseudopeople & Census, survey, tax, and administrative-style records & Stable simulant, household, employer, and linkage-truth identifiers & Multi-decade individual-based population dynamics & US entity-resolution research \\
Synthea & Synthetic patient records and interoperable EHR exports & Patient, provider, and organization entities & Forward lifespan and clinical-state simulation & Healthcare-specific longitudinal simulation \\
Mimesis 19 & Schema-generated structured records & Generic foreign-key references between schemas & No built-in longitudinal person-state process & Localized fake-data generation \\
SDV & Learned single-table, multi-table, and sequential synthetic data & Relationships and sequence keys supplied through metadata & Learned sequential structure when configured & Input-driven tabular synthesis \\
PANORAMA & Profile-grounded PII-rich web-style text & Sample-to-profile association; no released cross-person graph is central & No persistent state-update layer & Memorization and privacy corpus \\
PersonaBench & Timestamped private documents and QA tasks & Social graph used during generation; source profiles are not the benchmark interface & Timestamped sessions and selected updates & Personal-information retrieval benchmark \\
LoCoMo & Long multi-session dialogues and memory tasks & Dialogue participants and event grounding & Dated sessions and temporal event graphs & Long-term conversational memory \\
Privasis & Privacy-rich text with attribute annotations & Auxiliary profile conditioning; no reusable person graph is central & Text or document chronology rather than a person-state substrate & Privacy-rich text and sanitization \\
\bottomrule
\end{tabularx}
\caption{Closest adjacent resources, described by what users can inspect in the released artifact. Generation-only structures are not counted as released source objects. Internal consistency, statistical fidelity, human quality evaluation, leakage screening, provenance, and differential privacy are not collapsed into one ``audit'' mark.}
\label{tab:adjacent-resources}
\end{table*}

\begin{table*}[!htbp]
\centering
\scriptsize
\setlength{\tabcolsep}{3pt}
\begin{tabularx}{\textwidth}{@{}P{0.13\textwidth}P{0.29\textwidth}P{0.25\textwidth}Y@{}}
\toprule
Resource & Released artifact & Linking/time characteristics & Evidence and scope \\
\midrule
Privasis & Large privacy-rich text with personal-attribute annotations & Auxiliary profiles condition generation; no reusable cross-person source graph is the central release & Diversity, automated and human quality, overlap screening, and sanitization evaluation \\
PANORAMA & Profile-grounded PII-rich web-style text & Sample-to-profile grounding; no persistent person-state update layer & Memorization corpus and experiments \\
SynthPAI & Reddit-like interactions with verified personal-attribute labels & Synthetic profiles seed interacting agents; not a released longitudinal person-object population & Human realism checks and personal-attribute-inference experiments \\
PersonaBench & Timestamped private documents and personal QA tasks & Social graph and profiles support generation; evaluated models interact with benchmark documents rather than a reusable source graph & Retrieval-augmented personal-information QA \\
SPY & Synthetic PII-detection text with span labels & No cross-person graph or state process is central & PII detection data and benchmark evaluation \\
Nemotron-PII & Persona-grounded structured and unstructured records with PII spans & Persona context within examples; no linked population is central & PII/PHI detection; US and international conventions \\
Gretel finance PII & Multilingual financial text with PII annotations & Record-level examples; no person graph is central & Conformance and quality-oriented release scores \\
PIIBench & Unified multi-source PII detection corpus & Corpus unification rather than persistent people & Canonical PII types and evaluated baselines \\
LaMP & Personalized classification and generation tasks with user histories & Historical profile items and user/time splits; not a replayable state object & Personalization benchmark \\
LoCoMo & Long multi-session conversations and memory questions & Dated sessions and temporal event graphs & Human editing for consistency and long-term memory evaluation \\
LongMemEval & Multi-session memory tasks & Explicit temporal reasoning, knowledge updates, and abstention cases & Long-term assistant-memory benchmark \\
PersonaMem & Multi-session user interactions and personalized responses & Evolving user profiles and current-state evaluation & Dynamic profiling and personalization benchmark \\
PersonaChat & Persona-conditioned dialogue & Static persona statements; no persistent linked population & Dialogue consistency and engagement \\
Persona Hub & Large persona-description collection & Persona prompts rather than stable linked people & Conditioning source for synthetic data generation \\
Generative Agents & Interactive simulated agents with memories, reflection, and planning & Social interactions unfold over simulated time & Behavioral simulation and component ablations \\
\bottomrule
\end{tabularx}
{\footnotesize Sources: \citep{kim2026privasis,selvam2025panorama,yukhymenko2024synthetic,tan2025personabench,savkin2025spy,nvidia2025nemotronpii,gretel2024syntheticpii,jha2026piibench,salemi2024lamp,maharana2024evaluating,wu2025longmemeval,jiang2025personamem,zhang2018personalizing,ge2024personaHub,park2023generative}.\par}
\caption{Text, privacy, memory, persona, and behavior resources.}
\label{tab:comparison-text-memory}
\end{table*}

\begin{table*}[!htbp]
\centering
\scriptsize
\setlength{\tabcolsep}{3pt}
\begin{tabularx}{\textwidth}{@{}P{0.13\textwidth}P{0.29\textwidth}P{0.25\textwidth}Y@{}}
\toprule
Resource & Released artifact & Linking/time characteristics & Evidence and scope \\
\midrule
Pseudopeople & Census, survey, tax, and administrative-style tables & Stable simulant, household, and employer IDs across multi-decade dynamics; linkage truth provided & Census-scale US entity-resolution research \\
GeCo & Configurable personal-data generation and corruption & Generates linked/corrupted records for linkage experiments; not presented here as a general temporal population & Entity-resolution data generation \\
Febrl & Cleaning, standardization, deduplication, generation, and linkage toolkit & Record linkage and corruption workflows & Entity-resolution tooling \\
synthpop & Synthetic versions of input microdata & Preserves modeled relationships among input variables; no person-life process is required & Statistical disclosure control and microdata synthesis \\
SDV & Single-table, multi-table, and sequential synthesis & Metadata describes relationships and sequence keys & Input-driven learned synthesis and statistical-quality evaluation \\
PrivBayes & Differentially private tabular synthesis & Bayesian-network dependency structure; not a person-object graph & Formal DP mechanism and utility evaluation \\
PrivSyn & Differentially private tabular synthesis & Optimized marginal selection and data synthesis & Formal DP mechanism and utility evaluation \\
Synthea & Longitudinal synthetic EHRs and interoperable exports & Patient, provider, organization, disease, and care-process state over lifespans & Healthcare-domain simulation and interoperability \\
Faker & Localized fields and composite profiles & Values can be composed, but release-level cross-record commitments are not built in & General fake-data generation \\
Mimesis 19 & Localized fields and schema-generated records & Current schemas support foreign-key references between generated schemas; no built-in person-life process & General fake-data generation \\
\bottomrule
\end{tabularx}
{\footnotesize Sources: \citep{haddock2024pseudopeople,pseudopeople2026docs,tran2013geco,christen2008febrl,nowok2016synthpop,patki2016sdv,sdv2026docs,zhang2017privbayes,zhang2021privsyn,walonoski2018synthea,faker2025,mimesis2026schema}.\par}
\caption{Population, linkage, tabular, domain-simulation, and fake-data resources.}
\label{tab:comparison-population}
\end{table*}

\end{document}